\documentclass{stanford}

\usepackage{etoc}

\usepackage{microtype}
\usepackage{subcaption}
\usepackage{booktabs} 
\usepackage{dsfont}
\usepackage{bm, amsmath, amssymb, amsfonts, amsthm}
\usepackage{mathtools,commath}
\usepackage{graphicx, multirow}
\usepackage{xcolor}
\usepackage{hyperref}
\usepackage{tabularx}
\usepackage{natbib}

\usepackage[capitalize,noabbrev]{cleveref}

\theoremstyle{plain}
\newtheorem{theorem}{Theorem}[section]

\theoremstyle{definition}

\theoremstyle{remark}
\newtheorem{remark}[theorem]{Remark}

\usepackage[textsize=tiny]{todonotes}

\usepackage[most]{tcolorbox}   
\newtcolorbox{prompt}[1][]{
  colback=gray!5,          
  colframe=gray!80,        
  fonttitle=\bfseries,    
  title=Prompt,            
  left=2mm, right=2mm, top=2mm, bottom=2mm,
  boxrule=0.8pt,
  breakable,
  rounded corners,          
  #1                       
}

\tcbset{
    colback=gray!5!white, 
    colframe=black!75,    
    fonttitle=\bfseries,  
    boxrule=0.5mm,
    sharp corners=south   
}

\title{What LLMs Think When You Don’t Tell Them What to Think About?}

\author[1]{\text{Yongchan Kwon}}
\author[1,2]{\text{James Zou}}
\affiliation[1]{Together AI}
\affiliation[2]{Stanford University}

\dataset{\url{https://huggingface.co/datasets/ykwon-hf/unconditional_text_generation}}
\code{\url{https://github.com/ykwon0407/LLM_TOM}} 
\metadata[Correspondence to]{\texttt{jamesz@stanford.edu}}

\begin{document}
\onecolumn
\abstract{
Characterizing the behavior of large language models (LLMs) across diverse settings is critical for reliable monitoring and AI safety. However, most existing analyses rely on topic- or task-specific prompts, which can substantially limit what can be observed. In this work, we study what LLMs generate from minimal, topic-neutral inputs and probe their near-unconstrained generative behavior. Despite the absence of explicit topics, model outputs cover a broad semantic space, and surprisingly, each model family exhibits strong and systematic topical preferences. GPT-OSS predominantly generates programming (27.1\%) and mathematical content (24.6\%), whereas Llama most frequently generates literary content (9.1\%). DeepSeek often generates religious content, while Qwen frequently generates multiple-choice questions. Beyond topical preferences, we also observe differences in content specialization and depth: GPT-OSS often generates more technically advanced content (e.g., dynamic programming) compared with other models (e.g., basic Python). Furthermore, we find that the near-unconstrained generation often degenerates into repetitive phrases, revealing interesting behaviors unique to each model family. For instance, degenerate outputs from Llama include multiple URLs pointing to personal Facebook and Instagram accounts. We release the complete dataset of 256,000 samples from 16 LLMs, along with a reproducible codebase.}

\maketitle
\etocdepthtag.toc{mtchapter}

\section{Introduction}
\label{sec:intro}
Analyzing the behavior of large language models (LLMs) across diverse inputs and prompting conditions has become increasingly important for the safe and reliable deployment of AI systems \citep{liu2023trustworthy,rottger2025safetyprompts}. Behavioral auditing enables model developers to identify risks that are challenging to capture with standard benchmarks---such as the unintended generation of harmful content, biased outputs, or private information \citep{albuquerque2024evaluating}. At the same time, it helps end users make more informed decisions about which models to use. Accordingly, numerous studies have systematically investigated LLMs behaviors and biases \citep{tamkin2024clio,hu2025generative}. For instance, prior studies have examined inter-model homogeneity in open-ended settings~\citep{jiang2025artificial} and biased advice associated with names commonly linked to racial minorities and women~\citep{salinas2024s}. 

\begin{figure}[!h]
  \centering
  \begin{subfigure}[b]{0.48\textwidth}
    \includegraphics[width=\linewidth]{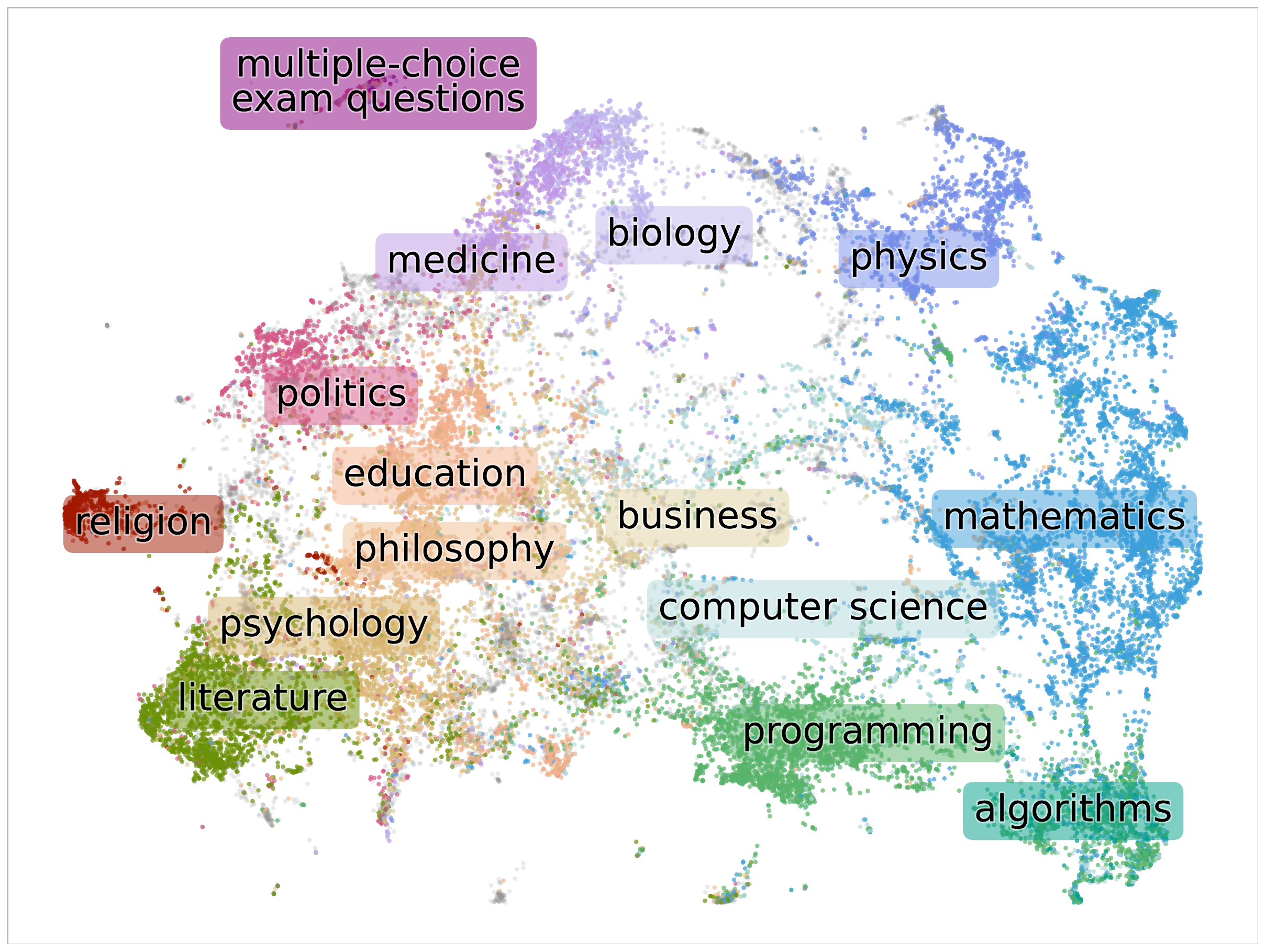} 
    \caption{}
    \label{fig:knowledge_space}
  \end{subfigure}
  \begin{subfigure}[b]{0.5\textwidth}
    \includegraphics[width=0.49\linewidth]{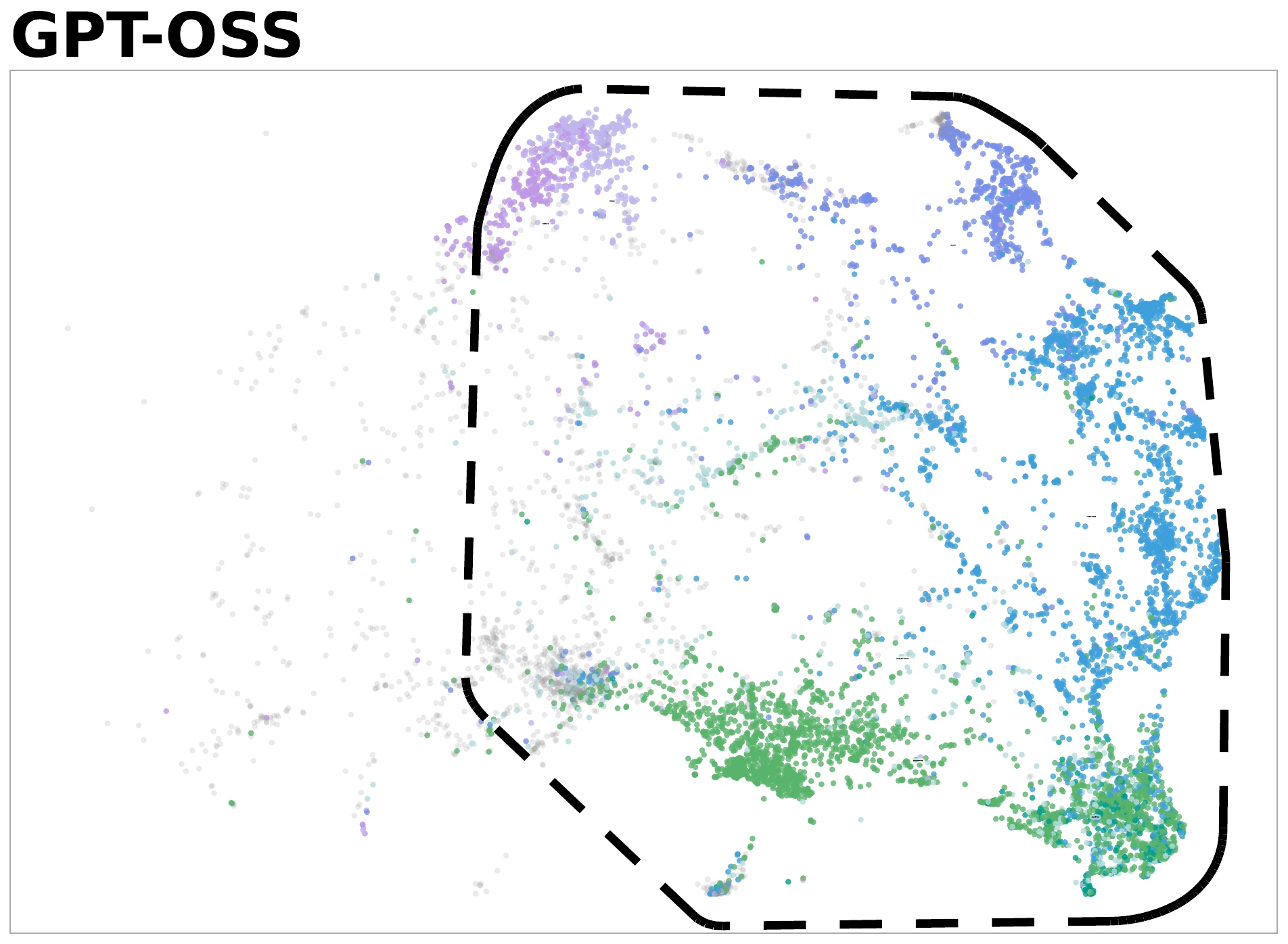}
    \includegraphics[width=0.49\linewidth]{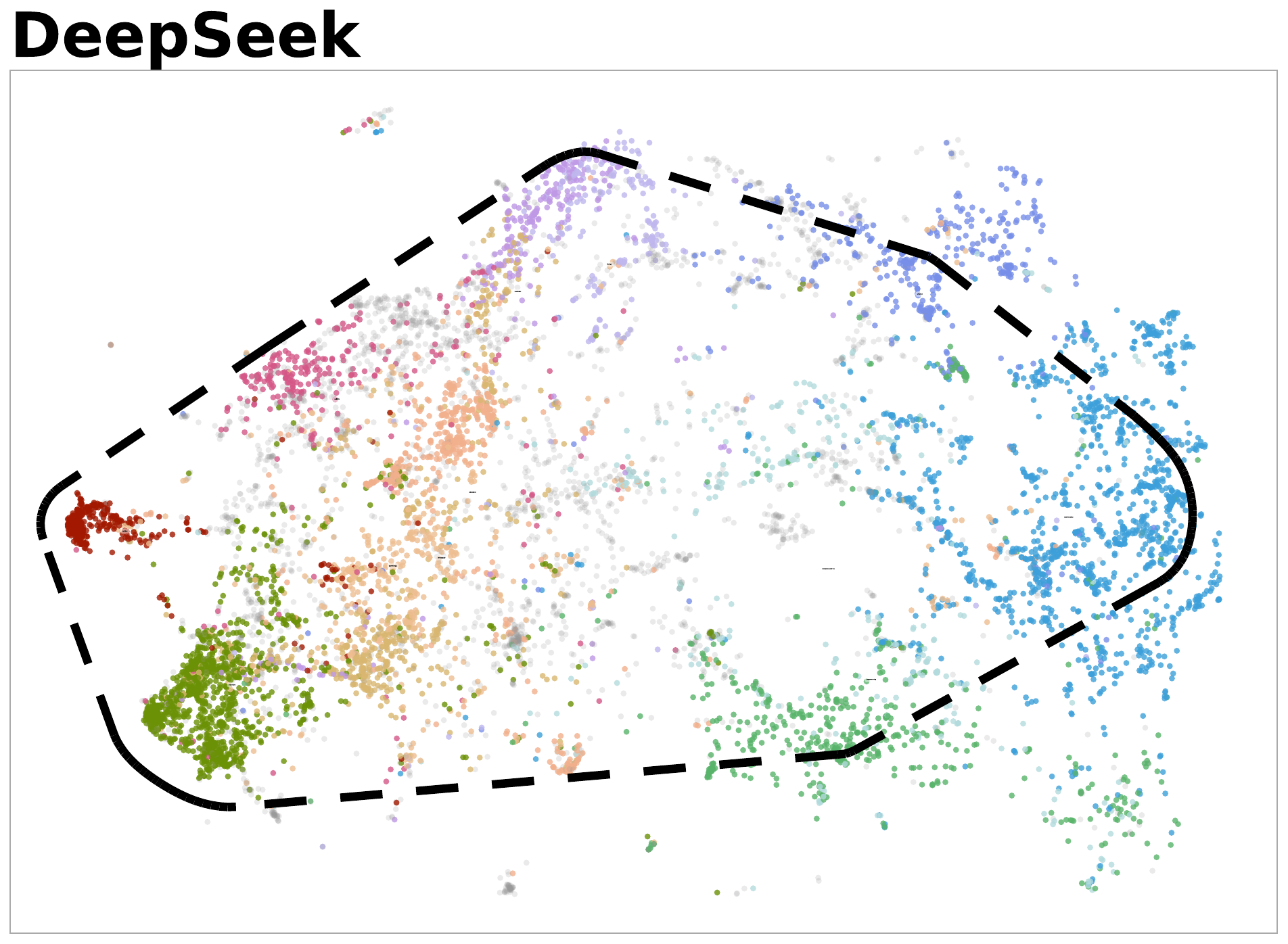}
    \\
    \includegraphics[width=0.49\linewidth]{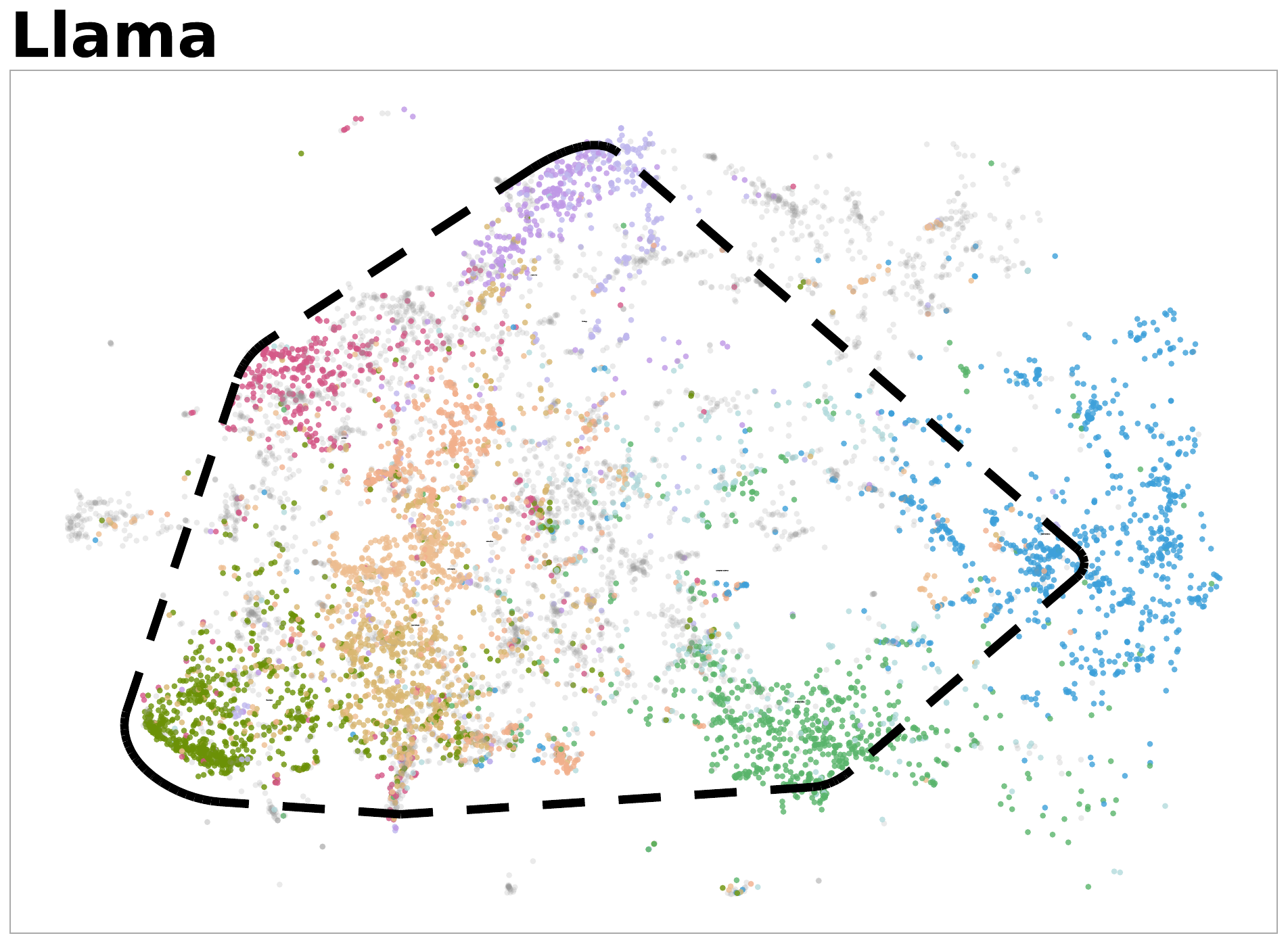}  
    \includegraphics[width=0.49\linewidth]{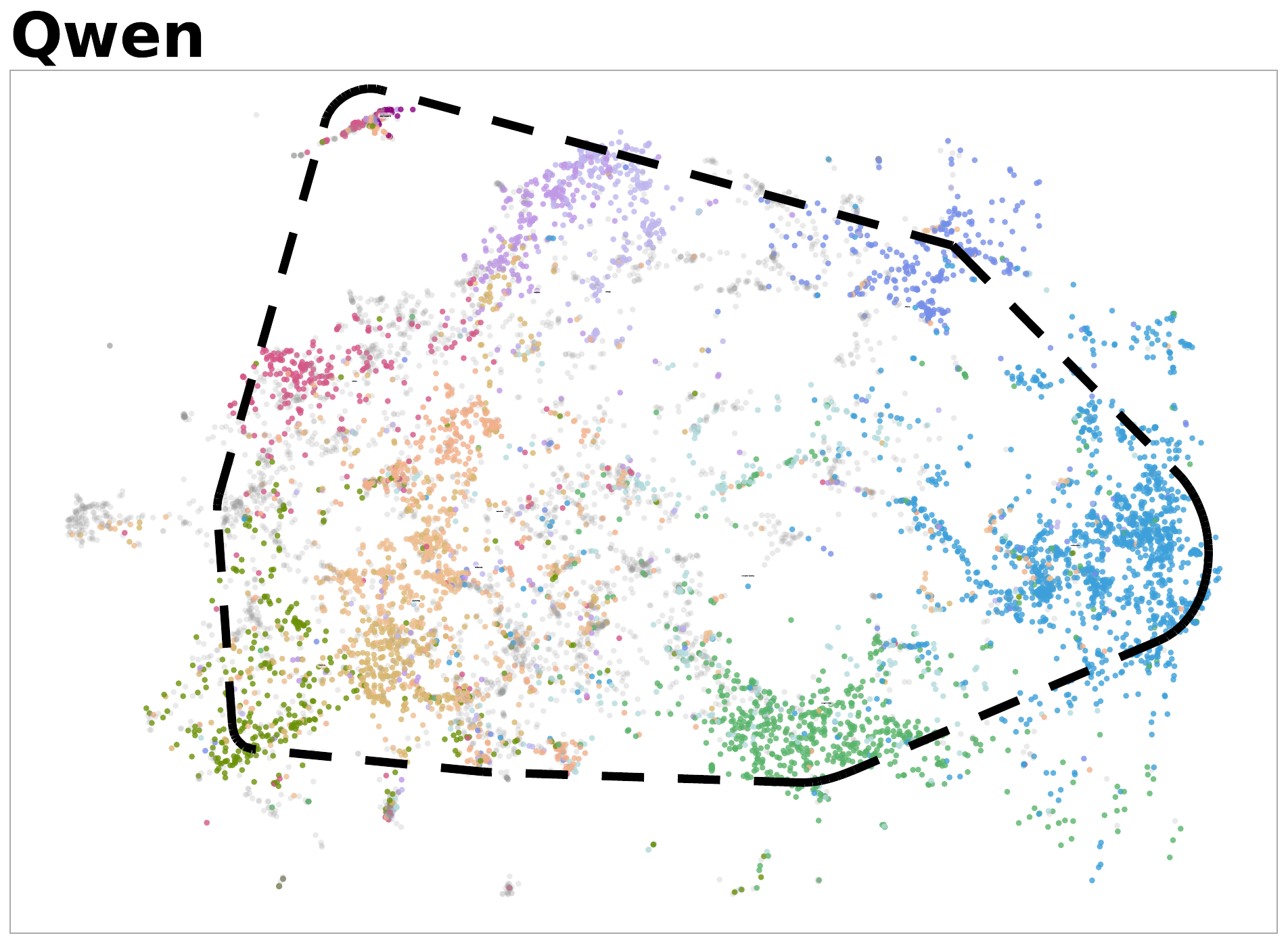}  
    \caption{}
    \label{fig:knowledge_space_individual}
  \end{subfigure}
  \\
  \begin{subfigure}[t]{\textwidth}
    \centering
    \includegraphics[width=\linewidth]{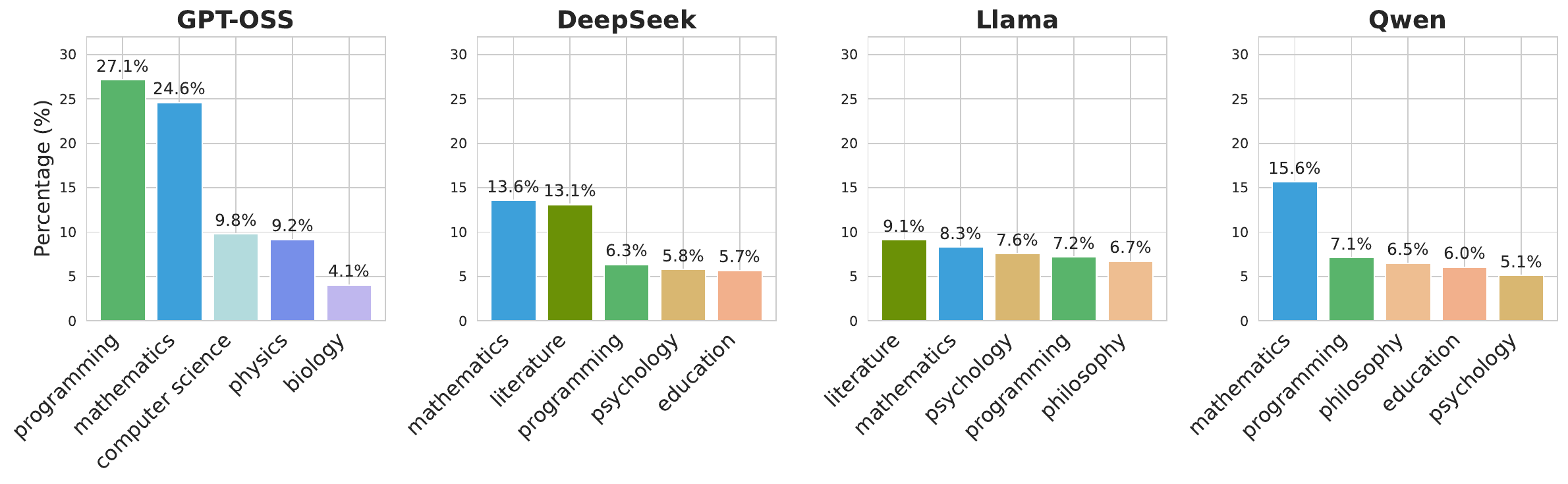} 
    \vspace{-0.2in}
    \caption{}
    \label{fig:category_charts}
  \end{subfigure}
   \caption{\textbf{LLM's top-of-mind behaviors.} UMAP visualizations of model outputs generated by the GPT-OSS, DeepSeek, Llama, and Qwen model families. Figure (a) visualizes outputs from all four model families, with each dot representing a generated output and class labels positioned at the centroid of their clusters. Except for ‘multiple-choice exam questions’ and ‘algorithms,’ cluster labels correspond to the 13 most frequent categories in our dataset. The two labels are included to better highlight clusters that are particularly prominent in Qwen and GPT-OSS. Figure (b) illustrates an individual model family, with the black dotted line indicating the convex hull of the high-density region. Figure (c) shows the top five categories within each model family and their corresponding percentages. The color scheme is consistent across all figures. \textbf{Despite the use of topic-neutral seed prompts, the semantic distributions of model outputs show a broad and diverse range of topics, and each model family exhibits distinctive distributional patterns.} Methodological details are provided in Section~\ref{sec:methods}, and a high-quality interactive figure with fine-grained labels is available in Supplementary Material.}
   \label{fig:embedding_visualization}
\end{figure}

However, much of the existing literature relies on narrowly scoped, topic- or task-specific prompts. While this paradigm has been effective for eliciting and measuring specific model behaviors and biases, it also introduces a fundamental limitation: such prompts strongly constrain LLM outputs, making it difficult to explore their generative behavior in unconstrained or minimally constrained settings~\citep{hida2024social, yun2025price}.
Independently, it is important to study how LLMs behave under minimally constrained settings, as these can reveal behaviors that are otherwise hidden by task-specific prompts or constrained input structures~\citep{carlini2021extracting, hoscilowicz2024large}. These hidden behaviors have the potential to provide a fundamental basis for stress-testing models and identifying potential risks that may arise during model development and deployment.

Motivated by these perspectives, we study LLM behavior under minimal, topic-neutral prompts, where external constraints are reduced as much as possible; we refer to this as the model’s ``top-of-mind" behavior. By minimizing the use of prompts and fixed chat templates that are typically applied by default, we focus on a more fundamental question: what types of content do models naturally generate when unconstrained, and how systematically do these tendencies vary across model families? This is the central scientific question of this paper.

\paragraph{Main Contributions} 
In this paper, we systematically explore what LLMs generate under minimally constrained settings across sixteen modern LLMs. Through large-scale generation, we find that model outputs span a broad semantic space, even in the absence of explicit topics (Figure~\ref{fig:knowledge_space}). Interestingly, each model family exhibits distinctive and systematic topical preferences (Figure~\ref{fig:knowledge_space_individual}). GPT-OSS mainly generates programming and math content, whereas Llama most frequently produces literary text. DeepSeek frequently produces religious content, while Qwen often generates multiple-choice questions. 
Our analysis also reveals substantial differences in subcategory specialization (Figure~\ref{fig:subcategories_math_programming}) and the technical depth across model families (Figure~\ref{fig:complexity_of_text}). For instance, GPT-OSS produces more technically advanced content than the other models do.

In addition, we treat degenerate text as a behavioral signal and find that its characteristic patterns vary significantly across model families (Figure~\ref{fig:degenerate_text}). In particular, GPT-OSS frequently produces formatting-related phrases (e.g., a code block  ``\textbackslash n\textbackslash n\verb|`|\verb|`|\verb|`|\textbackslash n\textbackslash n\verb|`|\verb|`|\verb|`|"), while Qwen often generates conversational artifacts (e.g., ``let me know'' or ``thank'') as well as Chinese text. Llama generates accessible URLs referencing personal social accounts (Figure~\ref{fig:degenerate_text_examples}). 

Together, our findings show that near-unconstrained generation not only reflects underlying model tendencies but also exposes risks relevant to AI reliability and safety, highlighting its practical utility and promise for monitoring model behavior. We release the complete dataset of 256,000 samples from sixteen modern LLMs, along with a reproducible codebase and an interactive figure.

\begin{table}[t]
\centering
\caption{\textbf{Examples of prompts and their style}. All the prompts are designed to be both \emph{topic-neutral} and \emph{open-ended}, while covering a wide range of scenarios, including the edge case of punctuation-only prompts. The complete list of seed prompts is provided in Appendix~\ref{app:prompts}.}
\label{tab:prompt_examples}
\begin{tabular}{ll}
\toprule
\textbf{Prompt Style} & \textbf{Example} \\
\midrule
Conversational softeners & ``\textsf{Actually,}'' \\
Chain of thoughts & ``\textsf{Let's break this down.}'' \\
Declarative prompts & ``\textsf{I want to talk about something.}'' \\
Rhetorical inquiries & ``\textsf{Shall we explore something?}'' \\
Informative expository prompts & ``\textsf{This article presents,}'' \\
Punctuation-only prompts & ``\textsf{.}'' \\
\bottomrule
\end{tabular}
\end{table}

\section{Method for Probing LLMs’ Top-of-Mind Behaviors}
\label{sec:methods}
Our method consists of three stages: (i) text generation, (ii) removal of degenerate text, and (iii) semantic labeling and embedding extraction for downstream data analysis. 

\paragraph{Text Generation}
Our generation setting has two key components: (C1) the use of \emph{topic-neutral, open-ended} seed prompts and (C2) the absence of chat templates.

For (C1), we carefully design a set of 36 \emph{topic-neutral, open-ended} seed prompts. Here, the term \emph{topic-neutral} indicates that the prompts do not explicitly specify any particular subject matter, and the term \emph{open-ended} indicates that the prompts are not targeted at a specific task and do not imply a unique correct answer, allowing models to continue freely. 

In addition, our prompt design emphasizes the following aspects: brevity to ensure minimal constraints, no named entities, no domain-specific content words, and diversity.
Our prompt set covers six different styles---conversational softeners, chain of thoughts, declarative prompts, rhetorical inquiries, informative expository prompts, and punctuation-only prompts---with six seed prompts for each style. It allows us to systematically probe a wide range of generative behaviors and ensures broad coverage across different scenarios. Examples of seed prompts are provided in Table~\ref{tab:prompt_examples}, and the complete list can be found in Appendix~\ref{app:prompts}.

For (C2), we do not use a chat template, which is commonly used in chat systems with a system prompt and role-based tags. Instead, the model receives only a randomly sampled seed prompt. This design choice reflects our experiment in Appendix~\ref{app:impact_of_chat_templates}, which shows that predefined chat templates can substantially steer generation behavior, often reducing output diversity and constraining the range of observable behaviors. 

With (C1) and (C2), we generate text autoregressively from a randomly selected seed prompt, continuing until either the maximum token limit is reached or an end-of-sequence token is sampled. For generation, we use standard hyperparameters: a temperature of 1.0 and a top-p of 0.9 throughout the paper. 

\begin{remark}[The impact of chat templates]
What happens when our seed prompt (e.g., ``\textsf{Let’s think step by step.}'') is used in conjunction with a standard chat template? In this setting, we observe that LLMs often attempt to clarify the user's intent with a short answer (e.g., ``\textsf{What would you like to think through step by step?}''), which stems from standard post-training procedures designed to accurately understand user intent. This suggests that standard generation settings based on fixed chat templates can substantially shape what models generate, underscoring the need for new approaches and frameworks to uncover behaviors that remain hidden under conventional prompting. 
\end{remark}

\paragraph{Removal of degenerate text}
While most text generated in the previous stage is coherent and meaningful, we observe that the removal of chat templates can lead LLM outputs to contain degenerate text---a consecutive substring that exhibits repetitive and semantically less coherent patterns \citep{holtzman2019curious}. To be more precise, we define a segment of model output as degenerate text if it meets all of the following criteria. First, it contains a consecutive sequence of characters, which we refer to in this paper as \emph{a repeated phrase}, that is at least 10 characters long. Second, this sequence is consecutively repeated at least five times. Third, the repeated portion constitutes at least 5\% of the entire output.
Because degenerate text generally contains little semantic information and, in our experiments, continues until it reaches the maximum number of tokens allowed, which accounts for more than 96\% of cases, we remove it by truncating the output from its first occurrence.

\begin{figure*}[t]
    \centering
    \includegraphics[width=\textwidth]{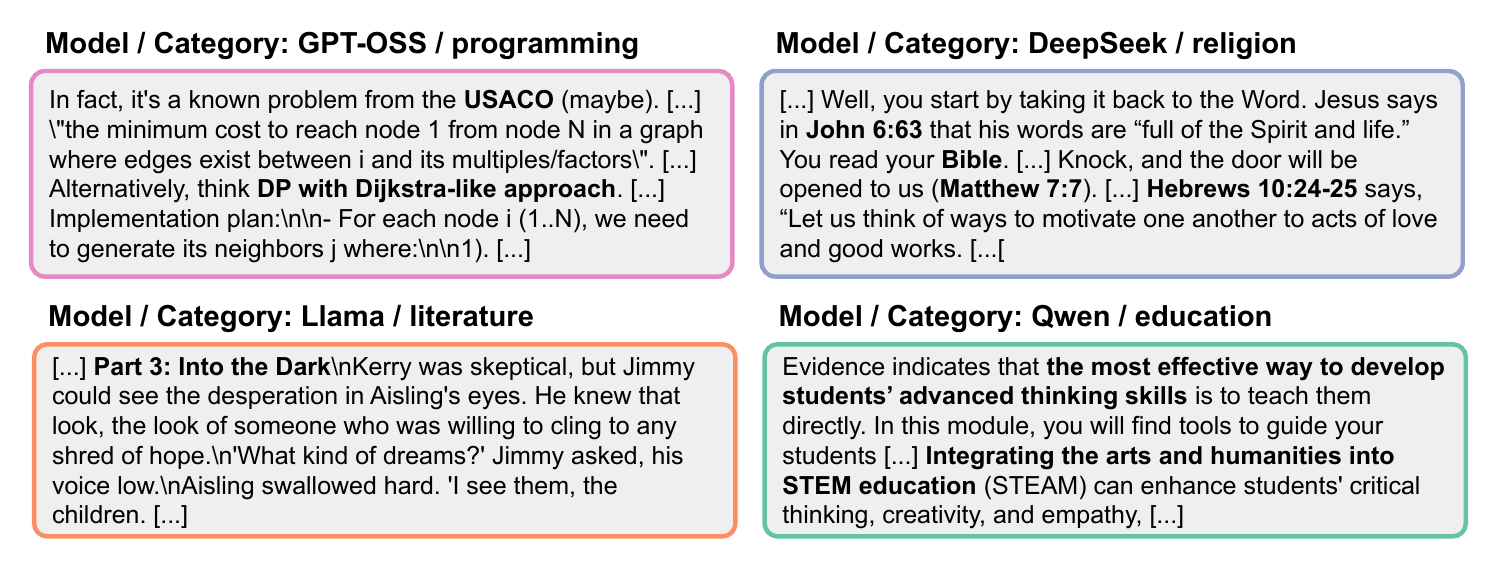}
    \vspace{-0.3in}
    \caption{\textbf{Examples of model outputs and their category labels}. We find that most generated texts appear coherent and meaningful, and that the LLM-based labeler produces sensible label annotations. For instance, the GPT-OSS sample describes an attempt to solve a problem that appears to be from the USA Computing Olympiad using a dynamic programming algorithm and is categorized as programming. The DeepSeek sample references Bible verses from Matthew and Hebrews and is categorized as religion. Llama generates a segment of fictional narrative, and Qwen outlines an approach to developing critical thinking; these outputs are accordingly categorized as literature and education, respectively.}
    \label{fig:generated_text_examples}
\end{figure*}

\paragraph{Semantic labeling and embedding extraction}
In the last stage of our method, we annotate semantic labels of the cleaned text from the previous stage and extract its embeddings for downstream data analysis. For labeling, following the practice in the literature \citep{gilardi2023chatgpt, tan2024large}, we adopt LLM-based open-vocabulary annotation. We instruct \texttt{GPT-OSS-120B} to annotate each sample with both a general category (e.g., mathematics) and a corresponding subcategory (e.g., topology). The prompts used for labeling are provided in Appendix~\ref{app:prompts}. For embedding extraction, we use \texttt{Qwen3-Embeddings-8B}, as it is among the most powerful open-source embedding models \citep{muennighoff2022mteb, enevoldsen2025mmtebmassivemultilingualtext}. We additionally conduct sensitivity analyses to examine the robustness of our results to the choice of LLM labelers (Appendix~\ref{app:sensitivity_analysis}) and embedding models (Appendix~\ref{app:sensitivity_analysis_on_embeddings}). These analyses show that our main findings are consistently observed across different settings, demonstrating the robustness of our results.

\section{Analysis of Model Behaviors under Topic-Neutral, Open-Ended Prompts}
\label{sec:llm_tom}
\subsection{Experimental settings}
We evaluate sixteen language models from four families, GPT-OSS, DeepSeek, Llama, and Qwen, for which the complete list of models is provided in Appendix~\ref{app:implementation_details}. Our selection covers a diverse range of model providers, reasoning styles (e.g., reasoning, hybrid, and standard instruct), and parameter scales from 3B to 671B parameters. We focus on open-source models to ensure that text generation is not influenced by any external system prompts, hidden instruction tuning, or post-processing potentially applied in API-based deployments \citep{chen2024chatgpt, gao2024model}.
For each model, we initially generate 16,000 samples with a maximum length of 4,096 tokens, yielding a total of 256,000 samples. We exclude 9,682 samples (3.8\% of the raw dataset) labeled as `unknown' or `failed' by the LLM-based annotator. This filtered dataset is used for all analyses in the paper. For visualization, we create a balanced subset by stratified sampling across model families, with 10,000 randomly selected samples per family. 

\begin{figure*}[t]
    \centering
    \includegraphics[width=0.95\linewidth]{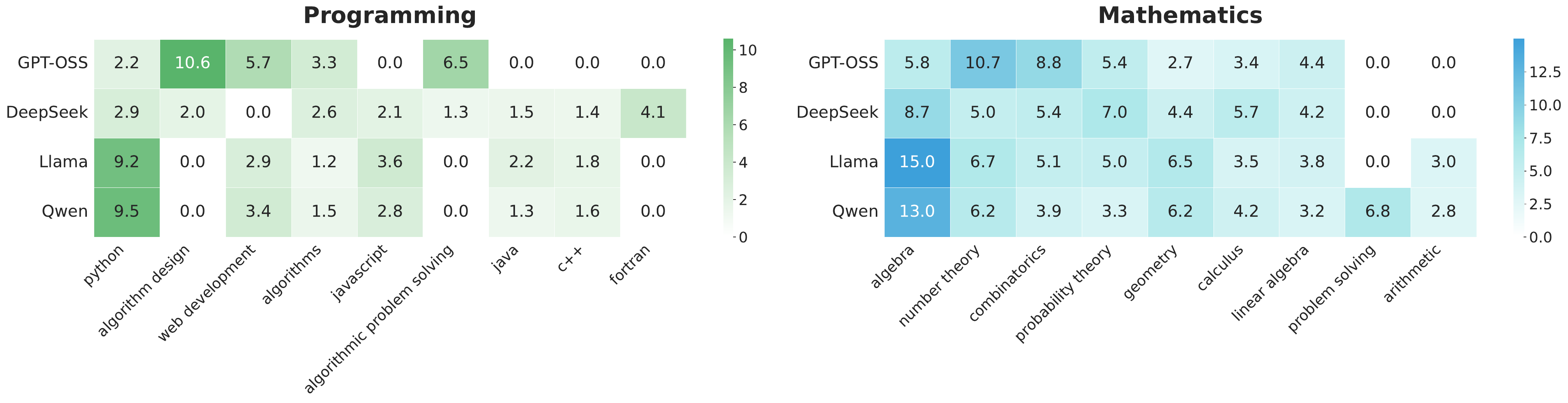}
    \caption{\textbf{Distribution of subcategories for programming and mathematics}. Numbers in each cell indicate the percentage of outputs that fall into a given subcategory among all outputs generated by the corresponding model family within a specific category. For each category, we select the nine subcategories with the highest average percentages and order them based on their size. \textbf{Each model family demonstrates distinct specialization across subcategories}.} 
    \label{fig:subcategories_math_programming}
\end{figure*}

\subsection{Main findings}
\paragraph{Broad and diverse topics even under topic-neutral prompts}
Figure~\ref{fig:knowledge_space} visualizes the semantic space of generated texts from all models \citep{mcinnes2018umap}. Despite the use of topic-neutral seed prompts, the generated outputs exhibit broad semantic coverage. We identify 123 well-populated representative categories, each containing at least 50 samples, which together account for 98.6\% of the entire dataset. These categories span a wide range of domains, including the liberal arts (e.g., literature, philosophy, and education), science and engineering (e.g., physics, mathematics, and programming), as well as areas such as law, finance, music, sports, cooking, agriculture, archaeology, military, and fashion. See Figure~\ref{fig:generated_text_examples} for representative examples.

\paragraph{Topical discrepancies across model families}
Interestingly, we observe that each model family exhibits distinct generative behaviors (Figure~\ref{fig:knowledge_space_individual}). In particular, as shown in Figure~\ref{fig:category_charts}, GPT-OSS mostly generates content related to programming (27.1\%) and mathematics (24.6\%), with these two categories alone accounting for over 50\% of its output. This combined proportion is substantially larger than that of any other model family; Qwen accounts for 22.7\% of the outputs, followed by DeepSeek at 19.9\%. In contrast, Llama generates the least technical content, but it produces diverse outputs in liberal arts domains, including literature (9.1\%), psychology (7.6\%), and philosophy (6.7\%). 
This concentration of GPT-OSS in scientific domains and Llama in liberal arts is further illustrated by their high-density regions in Figure~\ref{fig:knowledge_space_individual}, where kernel density estimation is used to estimate the embeddings’ density values.

DeepSeek and Qwen exhibit more balanced semantic distributions than the other two model families. Still, as shown in Figure~\ref{fig:knowledge_space_individual}, each exhibits distinct content preferences. DeepSeek generates religious content at a substantially higher rate (4.6\%) than other model families: Qwen (2.6\%), Llama (1.9\%), and GPT-OSS (less than 0.05\%). Similarly, Qwen forms a distinctive cluster dominated by multiple-choice exam questions, in which text often follows a structured format with a question and corresponding answer choices. We hypothesize that this prevalence reflects extensive use of multiple-choice exam data during Qwen’s training. All the results are consistent across different random splits of the prompts, supporting the robustness of our findings in Appendix~\ref{app:robustness_choice_prompts}.

\begin{remark}
One noteworthy observation is that, despite clear differences across model families, mathematics and programming consistently rank among the top five categories for all models. Given that LLM outputs tend to reflect patterns in their training data, this phenomenon may indicate a significant presence of mathematics- and programming-related content in training data, potentially driven by their importance and popularity in benchmark evaluations and AI applications \citep{cobbe2021training, jimenez2023swe}.
\end{remark}

\paragraph{Subcategory-level differences across model families} 
We further conduct a fine-grained analysis at the subcategory level. Specifically, we select the top nine subcategories in programming and mathematics, and examine differences in subcategory preferences across model families. Additional analyses of ten categories, including literature, psychology, physics, and chemistry, are provided in Appendix~\ref{app:fine_grained_analysis}. 

Figure~\ref{fig:subcategories_math_programming} illustrates distributions of subcategories for programming and mathematics. In programming, GPT-OSS most frequently generates content related to `algorithm design' and `algorithmic problem solving,' while the other model families generate content in general programming languages: Llama and Qwen primarily focus on `python,' and DeepSeek emphasizes `fortran.' In addition, the distribution in Figure~\ref{fig:subcategories_math_programming} includes many zero-valued entries, demonstrating that each model family specializes in distinct subcategories. 
In mathematics, GPT-OSS generates `number theory' content, whereas the other three model families most frequently produce content related to `algebra.' We observe that outputs categorized as `algebra' frequently involve only basic arithmetic or polynomial operations, indicating that all models except GPT-OSS tend to generate relatively basic mathematical content. 

\begin{figure}[t]
    \centering
    \includegraphics[width=0.75\columnwidth]{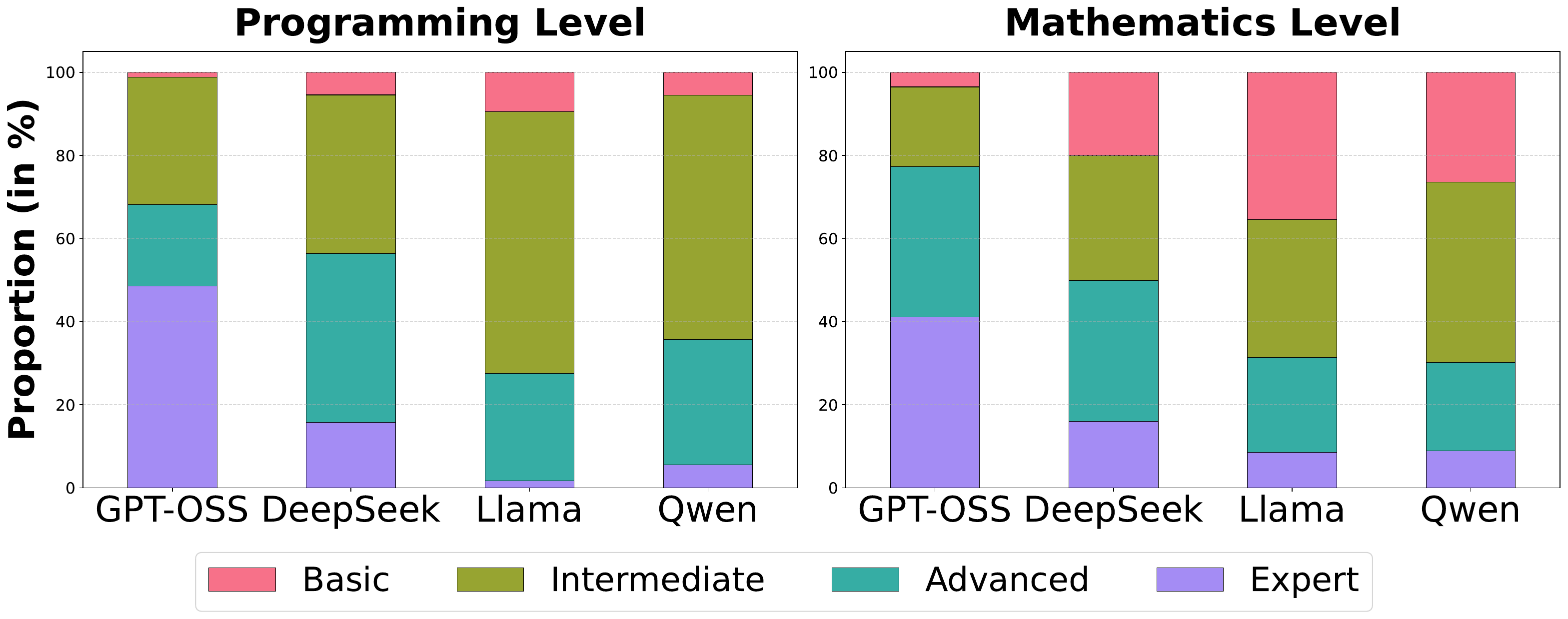}
    \caption{\textbf{Distribution of depth levels} in (left) programming and (right) mathematics. GPT-OSS mainly produces advanced or expert-level content in both programming and mathematics, while Llama and Qwen generate mostly basic or intermediate-level text.}
    \label{fig:complexity_of_text} 
\end{figure}

\begin{figure*}[t]
    \centering
    \includegraphics[width=\textwidth]{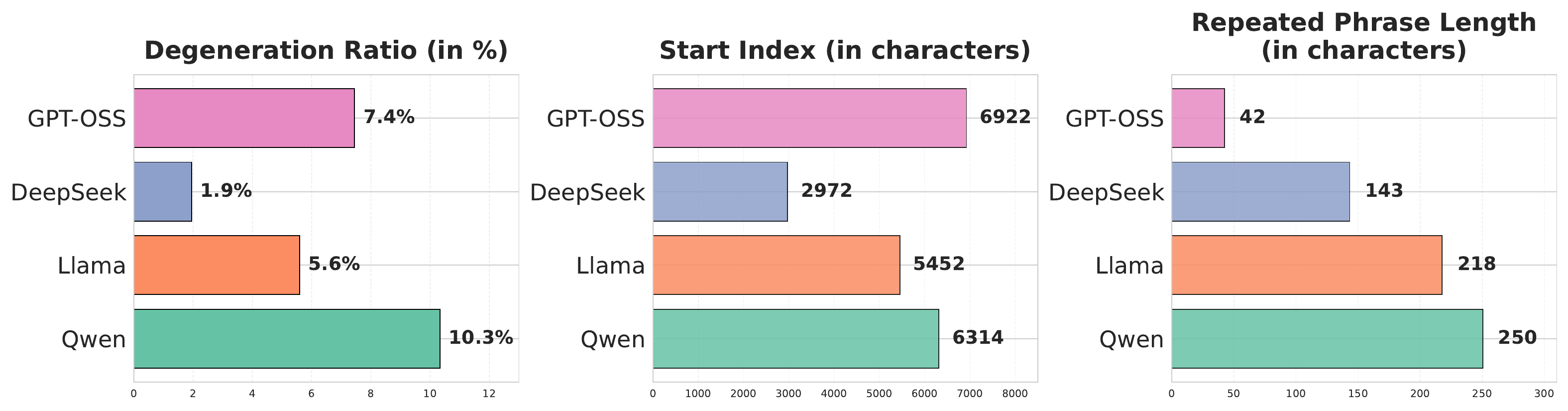}
    \caption{ \textbf{Distinct degenerate text behavior across model families.} The left figure shows how likely model outputs include degenerate text, the middle shows the start index of degenerate text within the generated sequence, and the right figure shows the average length of repeated phrases. We observe substantial variation across model families in the frequency of degenerate text, the position at which it begins, and the length of repetitive phrases. In particular, degenerate text occurs in only 1.9\% of DeepSeek generations, while it appears in 10.3\% of Qwen generations, representing a 5.4-fold increase.}
    \label{fig:degenerate_text}
\end{figure*}

\paragraph{Depth differences across model families}
Motivated by the preceding analysis, we analyze programming and mathematical text in terms of complexity. We focus on $n=31,732$ mathematics and programming samples, assigning each to one of four difficulty tiers---basic, intermediate, advanced, or expert---based on the detailed rubrics provided to \texttt{Claude‑4.5‑Opus}. These levels roughly correspond to elementary, middle, high school, and college-level content in mathematics, while in programming they cover a spectrum from basic concepts to competition- and interview-level questions. The exact prompt used in this experiment is provided in Appendix~\ref{app:prompts}.

As anticipated, GPT-OSS generates advanced and expert-level content more frequently than the other model families in both mathematics and programming (Figure~\ref{fig:complexity_of_text}). Specifically, 68.2\% of GPT-OSS’s programming outputs are classified as advanced or expert, primarily consisting of college-level algorithm questions (e.g., depth-first search, breadth-first search, or dynamic programming). In comparison, the corresponding percentages for DeepSeek, Qwen, and Llama are 56.4\%, 35.7\%, and 27.6\%, respectively. These findings indicate that differences across model families are reflected not only in their subcategory specialization but also in the complexity of the content they produce.

\begin{remark}[Topical preference and task competence]
The topical preferences and output complexity analyzed in this section capture a model’s knowledge space and its propensity to generate content related to particular domains. However, we find that these generative preferences do not reliably correspond to actual task competence. For example, GPT-OSS exhibits the lowest generation frequency in the law domain (less than 0.2\%) among the model families, yet its performance on law benchmarks is not necessarily inferior to that of other models \citep{guha2023legalbench}. We attribute this discrepancy to differences in generation settings, namely conditional versus near-unconditional generation. When prompted with explicit task context such as questions or instructions, model outputs are strongly guided by the prompt and can express detailed, task-specific knowledge. In contrast, when prompted with topic-neutral inputs without chat templates, model outputs reflect its naive probabilistic distribution acquired during training, which does not necessarily correspond to the model’s actual competence. 
\end{remark}

\section{Analysis of Degenerate Text}
\label{sec:degenerate}
Prior work \citep{welleck2019neural, holtzman2019curious} has shown that non-chat template settings frequently produce degenerate text, which is often discarded in downstream analyses due to low semantic coherence or excessive repetition. In this paper, rather than discarding it, we treat degenerate text as a behavioral signal and investigate its distributional patterns. 

Before analyzing inter-model differences, we first present the prevalence of this phenomenon at the population level. In our experiments in Section~\ref{sec:llm_tom}, we find that 6.8\% of all generations contain degenerate text. These samples frequently contain long repeated phrases, with an average length of 186 characters. This observation suggests that text degeneration is not the result of random decoding errors, but instead reflects a systematic component of model behavior. Below, we examine how these patterns vary structurally and distributionally across model families.

\begin{figure*}[t]
    \centering
    \includegraphics[width=\textwidth]{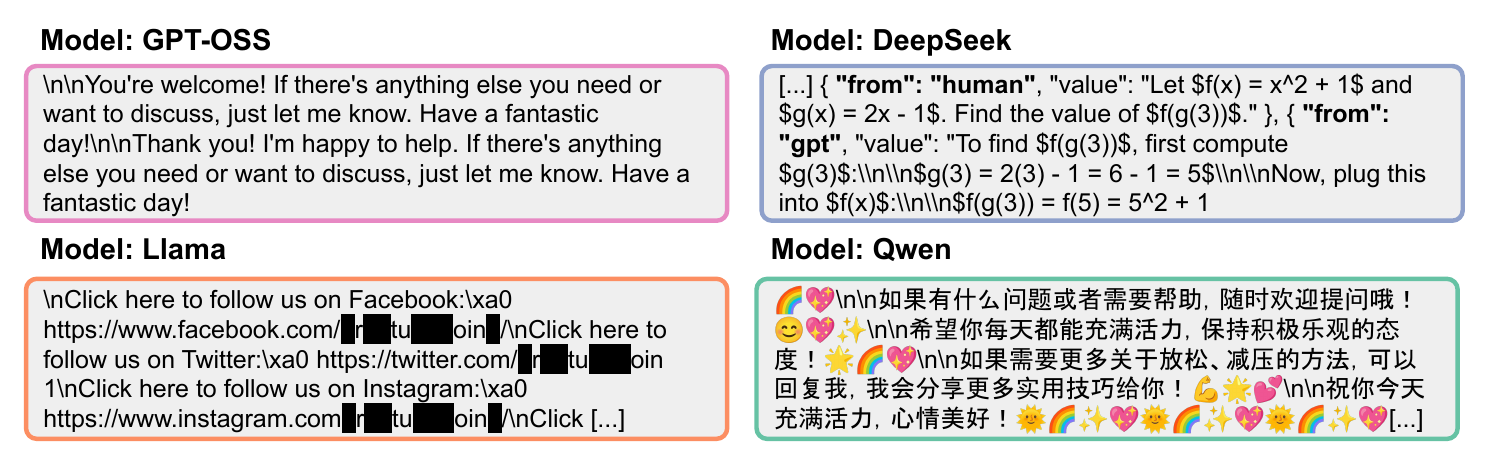}
    \caption{\textbf{Examples of repeated phrases in degenerate text.} We observe that degenerate text often exhibits distinct qualitative patterns across model families, including conversational artifacts in GPT-OSS and Chinese text in Qwen. Degenerate text reveals sensitive or personal information. In the Llama example, we mask portions of the URLs, as they are accessible to personal Facebook and Instagram accounts at the time of inspection. English translation of the Qwen sample is provided in Appendix~\ref{app:qwen_english}.}
    \label{fig:degenerate_text_examples}
\end{figure*}

\paragraph{Discrepancy in degenerate text across model families}
We first analyze (i) how frequently degenerate text occurs, (ii) when degeneration begins during generation, and (iii) the average length of repeated phrases. Figure~\ref{fig:degenerate_text} shows that model families exhibit distinct degeneracy behaviors. In terms of degenerate ratio, DeepSeek produces degenerate text in only 1.9\% of generations, whereas Qwen exhibits degeneration in 10.3\% of its outputs, representing a 5.4-fold difference. In terms of start index, degeneration in DeepSeek tends to occur earlier than other model families, with an average start index of 2,972 characters. In contrast, GPT-OSS and Qwen typically maintain coherent output for much longer, with degeneration beginning at 6,922 and 6,314 characters, respectively.

Regarding the length of a repeated phrase, Qwen exhibits the longest average sequence at 250 characters, followed by Llama with 218 characters, DeepSeek with 143 characters, and GPT-OSS with 42 characters. For GPT-OSS, this short length is largely attributable to formatting-related patterns. The top three phrases are a coding block ``\textbackslash n\textbackslash n\verb|`|\verb|`|\verb|`|\textbackslash n\textbackslash n\verb|`|\verb|`|\verb|`|'' (34.0\%), a truncation block ``\textbackslash n\textbackslash n...\textbackslash n\textbackslash n...'' (16.8\%), and an opening phrase ``\textbackslash n\textbackslash nBelow\textbackslash n\textbackslash nBelow'' (8.7\%), which together account for 59.5\% of all GPT-OSS phrase\footnote{The main reason the phrases have two identical sequences is that our criteria requires a minimum of 10 characters.}. These phrases are rare in other models appearing in only 0.1\% of Qwen outputs and absent in DeepSeek and Llama, and thus the phrases in other models are significantly longer than GPT-OSS. 

\paragraph{Artifacts in degenerate text}
We further examine repeated phrases and find that degenerate text frequently exhibits conversational artifacts, question-answer artifacts, and Chinese text. 

For conversational artifacts, many models generate polite or helpful phrases commonly used in user interactions, such as ``feel free to'', ``thank'', ``best regards'', and ``let me know''. These four expressions appear in 27.8\% of Qwen’s degenerate samples, 26.9\% for GPT-OSS, 14.6\% for Llama, and only 0.6\% for DeepSeek. Notably, although GPT-OSS and Qwen exhibit similar frequencies of such phrases, none of the GPT-OSS samples contain emojis, whereas Qwen frequently incorporates them (66.0\% of degenerate text with the conversational artifacts). Figure~\ref{fig:degenerate_text_examples} presents representative examples.

For question-answer artifacts, we identify repeated phrases associated with answer formatting, such as ``boxed'', ``final answer is'', and ``correct answer is''. These patterns occur in 6.1\% of Qwen’s degenerate samples, 4.5\% of Llama, 0.3\% of DeepSeek, and less than 0.01\% of GPT-OSS. For Chinese text, it is primarily observed in Qwen, representing 9.0\% of its degenerate samples, compared to 1.5\% for DeepSeek and less than 0.9\% for the other models. Taken together, these patterns highlight systematic distributional differences across model families.

\paragraph{A qualitative analysis of degenerate text}
Beyond model-family discrepancies, our analysis also surfaces several noteworthy qualitative phenomena, illustrated by the examples in Figure~\ref{fig:degenerate_text_examples}.
A sample from DeepSeek follows a structured JSON format, which is also known as ShareGPT format, with two entries. The first entry, labeled \textsf{``from": ``human"}, contains a mathematics question, while the second entry, labeled \textsf{``from": ``gpt"}, provides the corresponding solution. The presence of the second entry admits two possible explanations: (i) if the example was used during training, DeepSeek may be memorizing and reproducing it during generation. In this case, the example would constitute training data extraction from degenerate text, as studied by \citet{carlini2021extracting}. (ii) If it was not used during training, the model would instead be fabricating a plausible output without grounding in an actual source---a behavior that could be concerning, as it produces outputs that appear to imply a specific provenance (e.g., ChatGPT). 

Another example relates to the leakage of personal information. We find that degenerate text often includes specific URLs to personal Facebook, Instagram, or Twitter accounts. Specifically, using regular expression techniques to identify social accounts or email addresses, we estimate that Llama produces 0.78\% and Qwen 0.35\% of such instances. Fortunately, many of the generated links turned out to be invalid or hallucinated, but one sample generated by the Llama model included both Facebook and Instagram links that are actually accessible at the time of the investigation in January 2026. These examples show that monitoring degenerate text is crucial for model auditing, providing insights that are difficult to gain through conventional benchmark evaluations. We provide additional qualitative examples in Appendix~\ref{app:examples}. 

\section{Related Work}
\label{sec:related_work}

\paragraph{Behavioral analyses under topic-specific prompts}
Numerous prior studies have analyzed LLM behavior under topic-specific prompts or conversational settings, examining different aspects of model behavior, such as the diversity of model outputs \citep{santurkar2023whose, wu2024generative, murthy2025one, jiang2025artificial, xu2025echoes, hu2025generative, lu2026assistant}, political viewpoints expressed by models \citep{hartmann2023political, westwood2025measuring}, social biases manifested in generated content \citep{salinas2024s, wang2025large}, and the identification or mitigation of harmful behaviors through red-teaming efforts \citep{ganguli2022red, perez2022red}. While these studies provide invaluable insights under structured and purpose-built prompts, such settings inevitably shape model behavior and may not capture models’ near-unconstrained behaviors. Our work complements this line of research by analyzing models’ underlying behavior using minimal, topic-neutral generation settings.

\paragraph{LLM Fingerprints}
One practical implication of our findings is the presence of systematic inter-model discrepancies, which may be leveraged for LLM fingerprinting \citep{iourovitski2024hide, mcgovern2025your, gao2024model, bitton2025detecting}. Existing approaches, often statistical or feature-based, compare distributions of linguistic, syntactic, or embedding-level features across models \citep{hoscilowicz2024large, xu-etal-2024-instructional, bansal2025context, gao2024model}. Our results suggest that such distributional and stylistic patterns can naturally emerge under minimal, topic-neutral generation, indicating that they can complement traditional fingerprinting signals and provide a foundation for more robust and informative methods. 

\section{Discussion and Future Work}
\label{sec:discussion}
Our minimally constrained generation provides a useful and new lens for interpreting model behavior, but we should acknowledge several limitations of this work, which suggest directions for future research.

\begin{figure}[t]
    \centering
    \includegraphics[width=0.7\columnwidth]{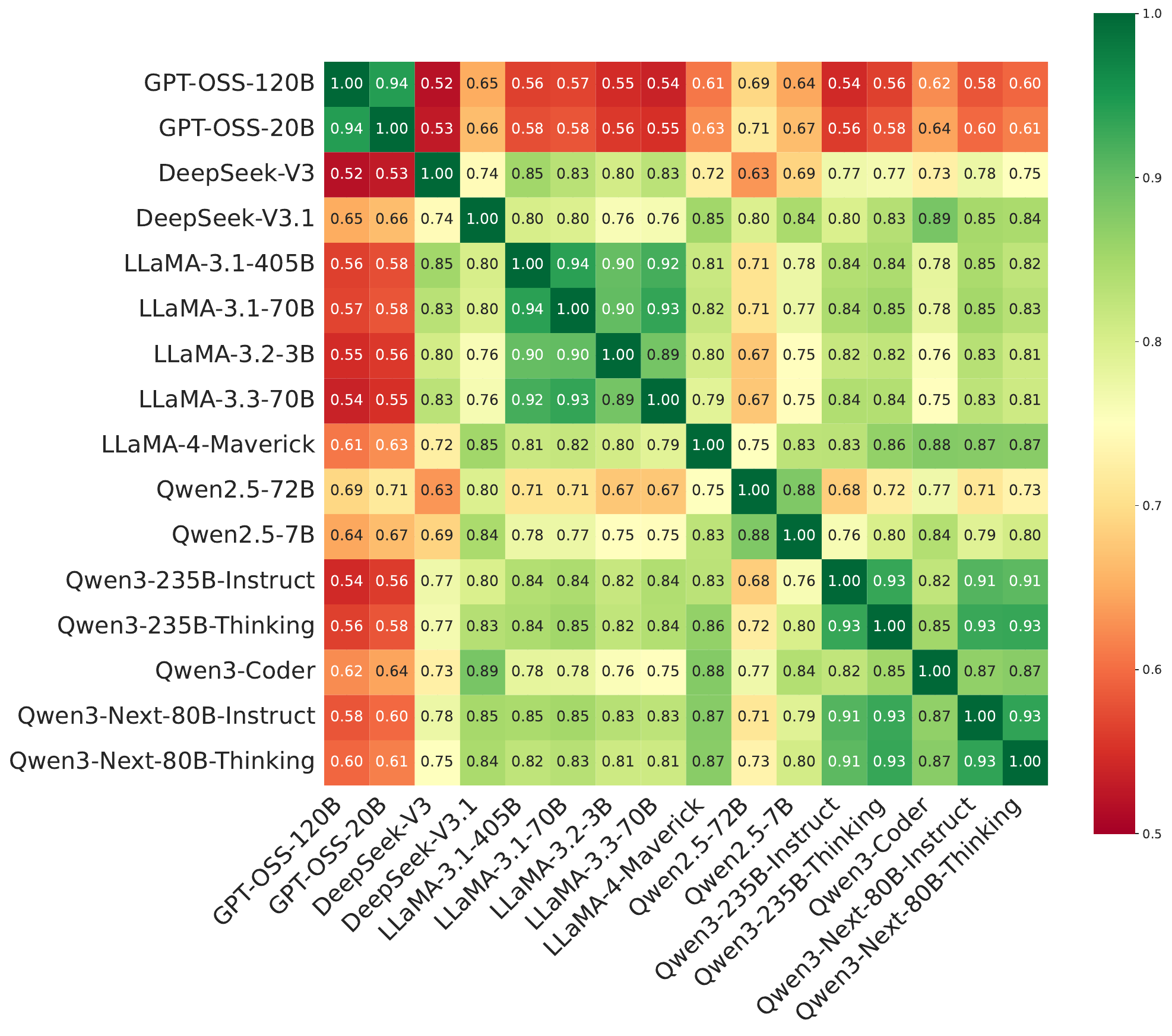}
    \caption{Similarity among the sixteen LLMs used in this paper. Similarity is measured as $1- \text{(JS divergence)}$ and ranges from 0 to 1, with higher values indicating greater similarity. Models from the same series but with different parameter sizes or reasoning styles generally exhibit high similarity, suggesting that distinguishing individual models within these groups is yet more challenging.}
    \label{fig:similarity_by_individual}
\end{figure}

\paragraph{Model-family level and individual-model level analysis}
While our analysis primarily focuses on differences at the model-family level, characterizing behavior at the level of individual models is important for many practical applications (e.g., LLM fingerprinting). However, this is inherently more challenging to detect. To quantitatively illustrate these challenges, we measure the similarity between every pair of models used in our experiments, using one minus the Jensen–Shannon (JS) divergence between their semantic label distributions.
Figure~\ref{fig:similarity_by_individual} illustrates these similarities across sixteen models. We find that, with the exception of DeepSeek, models within the same family exhibit high mutual similarity, and differences in parameter size or reasoning style result in only minor variations. In particular, the similarity between \texttt{GPT-OSS-120B} and \texttt{GPT-OSS-20B} is 0.94, and between \texttt{Qwen3-235B-Instruct} and \texttt{Qwen3-235B-Thinking} is 0.93. This suggests that distinguishing individual models based on category information can be challenging and likely requires additional signals or finer-grained analyses, highlighting an important direction for future work. In Appendix~\ref{app:classification_individual_model}, we further discuss this challenge of distinguishing individual models using embeddings from the lens of LLM fingerprints.

\paragraph{Prompt design and its impact} 
All prompts in our study are intentionally designed to be topic-neutral and open-ended. To avoid over-reliance on any single prompt formulation, we consider a diverse collection of seed prompts spanning multiple prompt styles, and further examine how prompt choice influences our main findings (Appendix~\ref{app:robustness_choice_prompts}). However, we also find that some variation emerges across different prompt styles, reflecting differences in phrasing and presentation (Appendix~\ref{app:effect_of_prompt_style}). These observations suggest that prompt style can shape the resulting label distributions, but that our use of diverse prompt formulations reduces the likelihood that the main patterns we report are driven by a narrow or idiosyncratic choice of prompts. We believe exploring a broader space of prompt designs remains an interesting direction for future work.

\section{Conclusion}
\label{sec:conclusion}
We systematically study how modern LLMs behave under unconstrained generation settings and show that, even under minimal prompting, model outputs span a broad semantic space while exhibiting distinct, family-specific topical preferences and levels of technical depth. We further identify significantly different patterns of degenerate text across model families, and our qualitative analysis reveals potential safety and privacy risks. Overall, our results demonstrate that minimally conditioned generation is an effective lens for monitoring underlying model tendencies. 

\section*{Impact Statement}
Our dataset is mainly produced by LLMs, it may include text content that is potentially harmful, offensive, or socially biased (e.g., stereotypes or derogatory language). While we manually inspected a subset of the dataset and applied exact matching-based filters to remove sensitive words, not all samples were reviewed by human. Therefore, despite these efforts, the released data may still include undesirable content. To reduce risk, we will apply conservative redaction and content filtering before and after release, and we will provide clear documentation describing known limitations of the dataset and recommended safe-use guidelines (e.g., avoiding deployment in settings that may expose end users to unmoderated outputs). 


\bibliographystyle{unsrtnat}
\bibliography{ref}

\newpage
\appendix

\section*{Appendix}
In this Appendix, we present additional experiments (Appendix~\ref{app:additional_experiments}), implementation details (Appendix~\ref{app:implementation_details}), prompts (Appendix~\ref{app:prompts}), and illustrative examples (Appendix~\ref{app:examples}).




\section{Additional Experiments}
\label{app:additional_experiments}

\subsection{Impact of Chat Templates}
\label{app:impact_of_chat_templates}

\citet{jiang2025artificial} shows inter-model homogeneity, where different models converge on similar ideas in open-ended settings. Here, we reproduce their experimental setup and additionally evaluate an ablated variant in which chat templates are removed. Given that the two settings differ only in the use of chat templates, their inputs in terms of context information are essentially the same; therefore, we would anticipate comparable behavior. However, our results reveal a qualitatively opposite trend in inter-model diversity.

\begin{figure*}[h]
    \centering
    \includegraphics[width=0.95\textwidth]{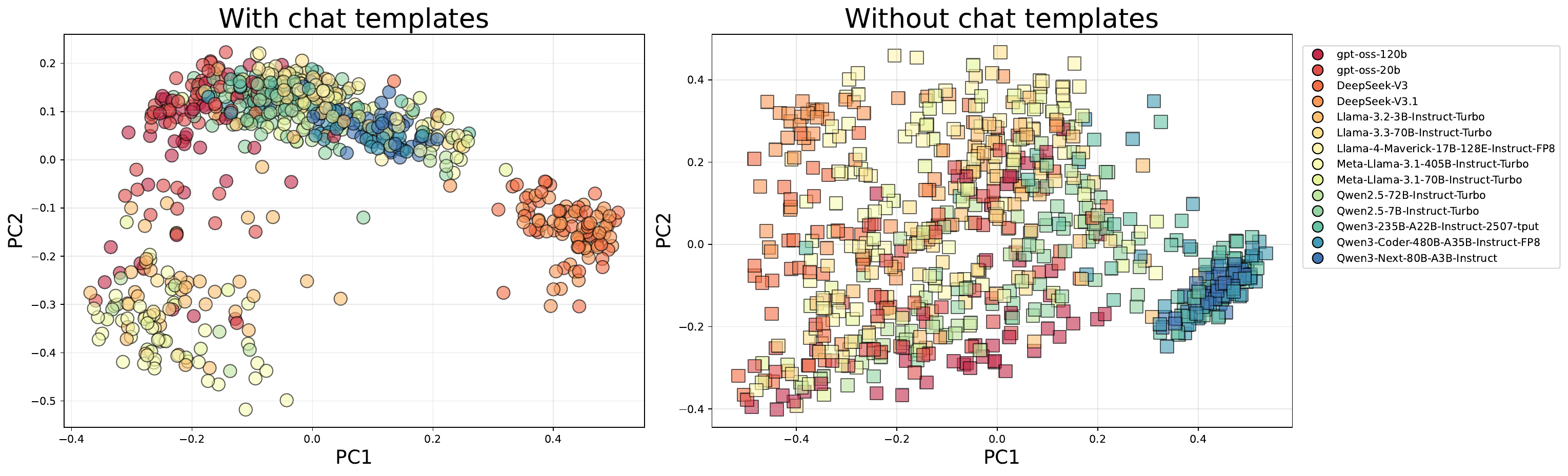}
    \caption{\textbf{Effect of removing chat templates on generative distributions}. The first two principal components of text generations from fourteen models in the legend, evaluated (left) with and (right) without chat templates. For each model, we generate 50 independent samples using the prompt ``\textit{Write a metaphor about time.}'' with a temperature of 1.0 and a top-p of 0.9, following \citet{jiang2025artificial}. Removing chat templates leads to more widely dispersed PCA representations and substantially more diverse metaphors. This indicates a template-induced shift in the models’ generative distributions.}
    \label{fig:chat_template_comparison}
\end{figure*}

Figure~\ref{fig:chat_template_comparison} illustrates how fixed chat templates constrain model behavior. When prompted with ``Write a metaphor about time.'' the standard chat template yields outputs that largely collapse into two dominant themes: ``time as a river'' and ``time is a weaver.'', as reported in the previous work \citep{jiang2025artificial}. However, the figure shows that removing the chat template produces more widely dispersed PCA representations and substantially more diverse metaphors (e.g., time as a ``mirror", ``architect", ``currency", or ``shadow"). This suggests that removing chat templates is more effective for exploring the diverse behaviors of models, echoing a previous work by \citet{yun2025price}. Accordingly, we do not apply chat templates in our experiments and investigate LLMs' behaviors under near-unconditional settings. 

\subsection{Sensitivity Analysis on LLM-based Labeling}
\label{app:sensitivity_analysis}

\paragraph{Motivation for Selecting the LLM Labeler}
Throughout our experiments, we used \texttt{GPT-OSS-120B} to assign category labels to model outputs. This choice was motivated by three observations. First, we initially experimented with n-gram–based keyword extraction methods \citep{campos2020yake, grootendorst2020keybert}; however, we found that these approaches are ineffective for labeling mathematical proofs and programming code. Second, based on approximately 100 manually labeled samples, we observed that human annotation is both costly and insufficiently reliable, as LLM outputs span an exceptionally broad range of topics (Figure~\ref{fig:knowledge_space}). Third, we found that this labeling task does not require highly complex models; at the same time, \texttt{GPT-OSS-120B} provides strong efficiency and instruction-following capabilities, which are critical for producing large-scale, well-formatted JSON outputs. At the very beginning of this exploration, we conducted a rough inspection of approximately 200 labels produced by \texttt{GPT-OSS-120B} and found them to be sufficiently reliable, although we did not perform a precise quantitative evaluation.

\begin{figure}[h]
    \centering
    \includegraphics[width=0.8\linewidth]{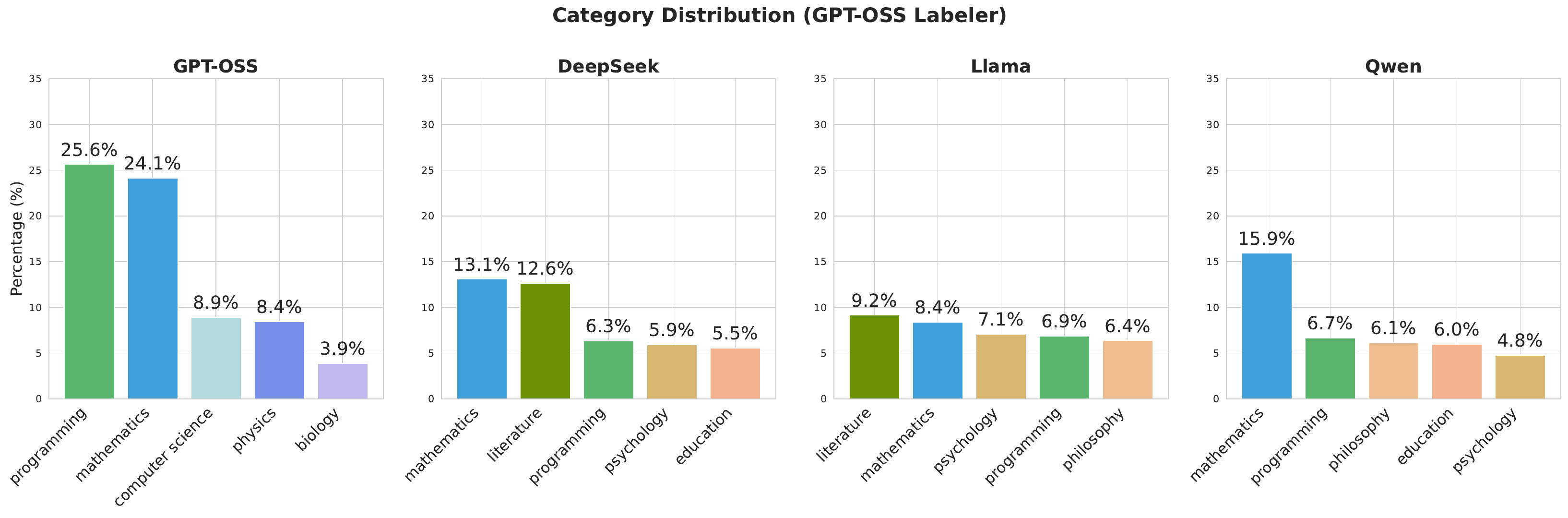}
    \includegraphics[width=0.8\linewidth]{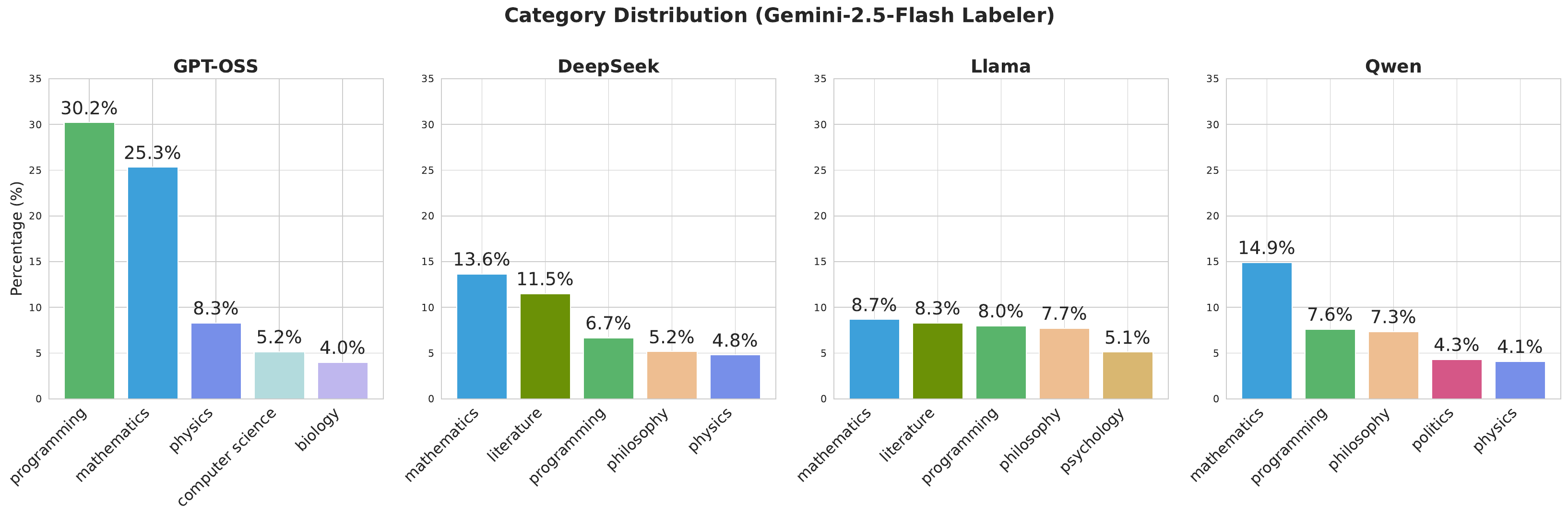}
    \includegraphics[width=0.8\linewidth]{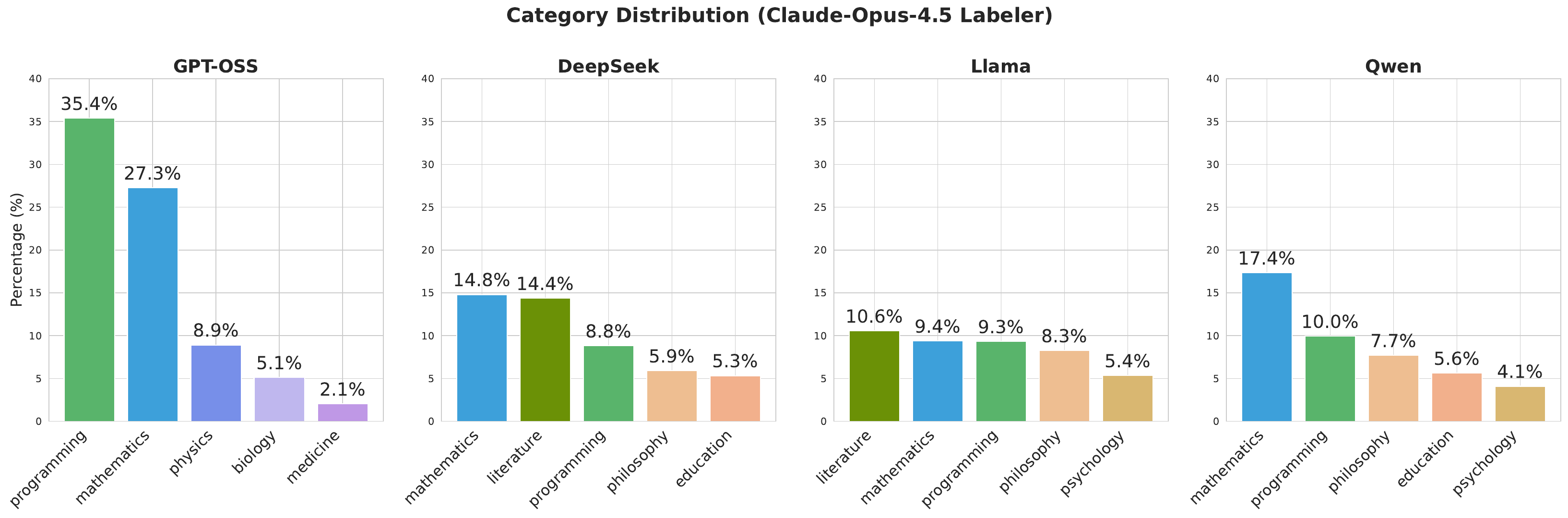}
    \caption{Sensitivity analysis of the category distributions. (Top) \texttt{GPT-OSS-120B}, (Middle) \texttt{Gemini-2.5-Pro-Flash}, and (Bottom) \texttt{Claude-4.5-Opus}. The analysis is conducted on a subset of $32,768$ model generations; consequently, the counts for \texttt{GPT-OSS-120B} differ slightly from those in Figure~\ref{fig:category_charts}. Overall, consistent with our main findings, each model family exhibits distinctive distributional patterns, and the general trends are stable regardless of the LLM used for labeling.}
    \label{fig:sensitivity_analyis_labelers}
\end{figure}

\paragraph{Analysis of Potential Dual Bias}
A natural concern is whether the dual use of \texttt{GPT-OSS-120B} for both text generation and semantic labeling could introduce bias into our conclusions. To examine this possibility, we conduct a sensitivity analysis. Specifically, we uniformly sample $32,768$ model generations from the full dataset, which accounts for 12.8\% of the full dataset, and compare the original labels produced by \texttt{GPT-OSS-120B} with those generated by \texttt{Gemini-2.5-Pro-Flash} and \texttt{Claude-4.5-Opus}. Figure~\ref{fig:sensitivity_analyis_labelers} reproduces the category distributions for this subset across the three LLM labelers.

Across all three LLM labelers, the top five categories are largely consistent across labelers, suggesting the robustness of LLM-based semantic annotation. In particular, we observe consistent patterns in the proportion of math- and programming-related outputs (Figure~\ref{fig:sensitivity_analyis_labelers}). All three labelers indicate that GPT-OSS frequently generates mathematical or programming-related content, with the combined proportion exceeding 50\% across labelers, which is observed in Figure~\ref{fig:category_charts}. Moreover, Llama places relatively less emphasis on science and technology, which is also observed in Figure~\ref{fig:category_charts}, as evidenced by having the lowest combined proportion of mathematics and programming outputs across all LLM labelers. Although absolute proportions vary slightly due to annotation differences, the relative ordering of categories and overall trends in major categories remain stable. This consistency suggests that any potential bias arising from the dual use of \texttt{GPT-OSS-120B} is not strong enough to affect our main findings. Our conclusions based on \texttt{GPT-OSS-120B} are robust to the choice of LLM used for semantic labeling; accordingly, we use \texttt{GPT-OSS-120B} to label the entire dataset.

\begin{table}[h]
\centering
\caption{Agreement between LLM-based labelers.}
\label{tab:labeling_robustness}
\begin{tabular}{lcc}
\toprule
\textbf{Comparison} & \textbf{Exact Match} \\
\midrule
\texttt{GPT-OSS-120B} vs.\ \texttt{Claude-4.5-Opus}     & 74.97\% \\
\texttt{GPT-OSS-120B} vs.\ \texttt{Gemini-2.5-Flash}     & 71.72\% \\
\texttt{Claude-4.5-Opus} vs.\ \texttt{Gemini-2.5-Flash}    & 71.54\% \\
\midrule
All Three Match      & 63.29\% \\
\bottomrule
\end{tabular}
\end{table}

Table~\ref{tab:labeling_robustness} further reports the agreement across labelers, measured using exact match\footnote{We find that strict exact matching does not fully capture agreement between models, as they often express the same concept using slightly different forms (e.g., (``sport", ``sports") and (``social science", ``social sciences")). We therefore manually merge such cases and report a relaxed exact match metric. With the strict exact match, the agreement was approximately 3\% lower.}. Overall, we observe substantial consistency across labelers, suggesting again that our results are not overly sensitive to the choice of labeling model. The main disagreement pairs are (medicine, biology) and (computer science, mathematics), where these are also reasonable because a model generation can contain multiple topics (e.g., writing a Python code for mathematical problems). We expect that incorporating majority voting from multiple independent annotations could further reduce labeling bias; however, we also believe our main findings would be consistent as the general tendency is similar across model families in Figure~\ref{fig:category_barplot_robustness_split_half}.

\begin{figure}[h]
    \centering
    \includegraphics[width=0.7\linewidth]{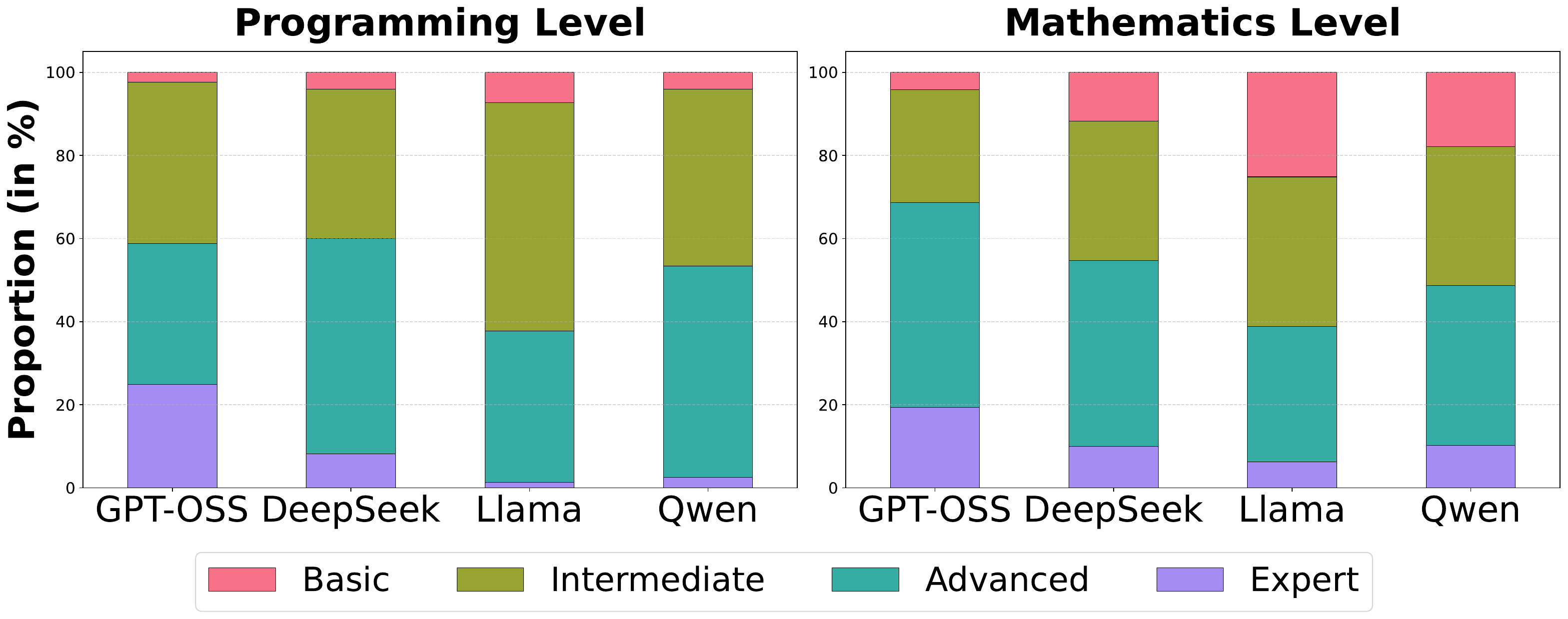}
    \caption{Distribution of depth levels for \texttt{Gemini-2.5-Flash}. Consistent with the results for \texttt{Claude-4.5-Opus} shown in Figure~\ref{fig:complexity_of_text}, GPT-OSS generates a larger proportion of text at the Advanced or Expert level.}
    \label{fig:programming_math_level_by_models_gemini}
\end{figure}

Lastly, we examine the robustness of our results across different LLMs in Figure~\ref{fig:complexity_of_text}. Specifically, we consider \texttt{Gemini-2.5-Flash} and relabel the complexity of outputs categorized as mathematics and programming. Figure~\ref{fig:programming_math_level_by_models_gemini} shows the proportions across the four depth levels. While the differences are less pronounced than those in Figure~\ref{fig:complexity_of_text}, GPT-OSS still demonstrates a concentration at the Advanced and Expert levels, indicating that our findings are robust.

\subsection{Sensitivity Analysis on Embedding Model}
\label{app:sensitivity_analysis_on_embeddings}
Following the sensitivity analysis in Appendix~\ref{app:sensitivity_analysis}, a natural next question concerns the robustness of our visualization to the choice of embedding model. In this subsection, we examine whether our visualization remains stable across different embedding models. Specifically, we consider an alternative open-source, competitive embedding model, \texttt{Llama-embed-nemotron-8b} \citep{babakhin2025llamaembednemotron8buniversaltextembedding} and reproduce UMAP visualization plots in Figure~\ref{fig:embedding_visualization}. The implementation settings are identical to those used in the previous analysis with \texttt{Qwen3-Embedding-8B}, except that the UMAP hyperparameter ``n\_neighbors" is set to 8 instead of 5 to improve visual comparability.

\begin{figure*}[h]
  \centering
  \begin{subfigure}[b]{0.48\textwidth}
    \includegraphics[width=\linewidth]{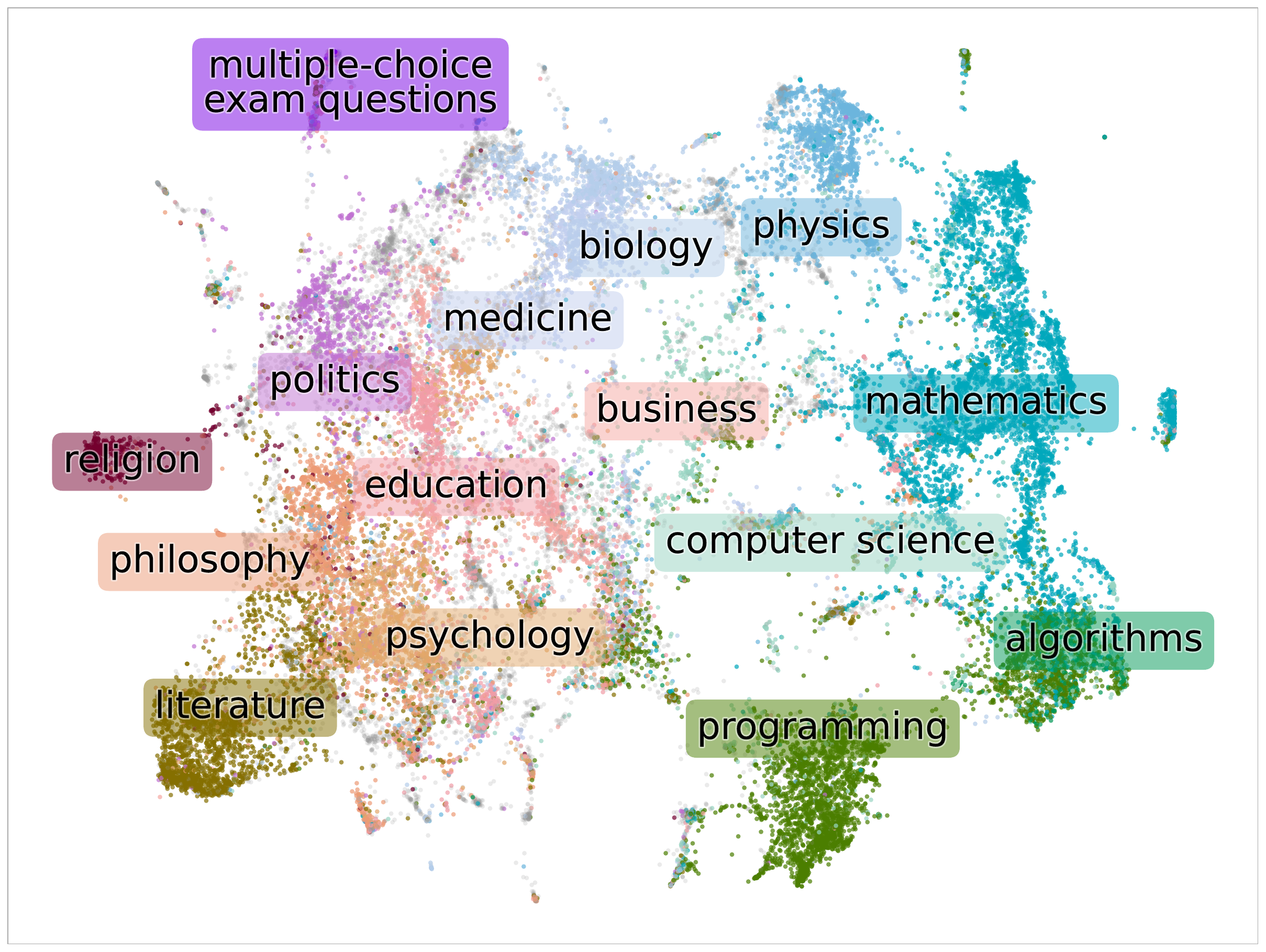} 
    \caption{}
    \label{fig:knowledge_space_nemotron}
  \end{subfigure}
  \begin{subfigure}[b]{0.5\textwidth}
    \includegraphics[width=0.49\linewidth]{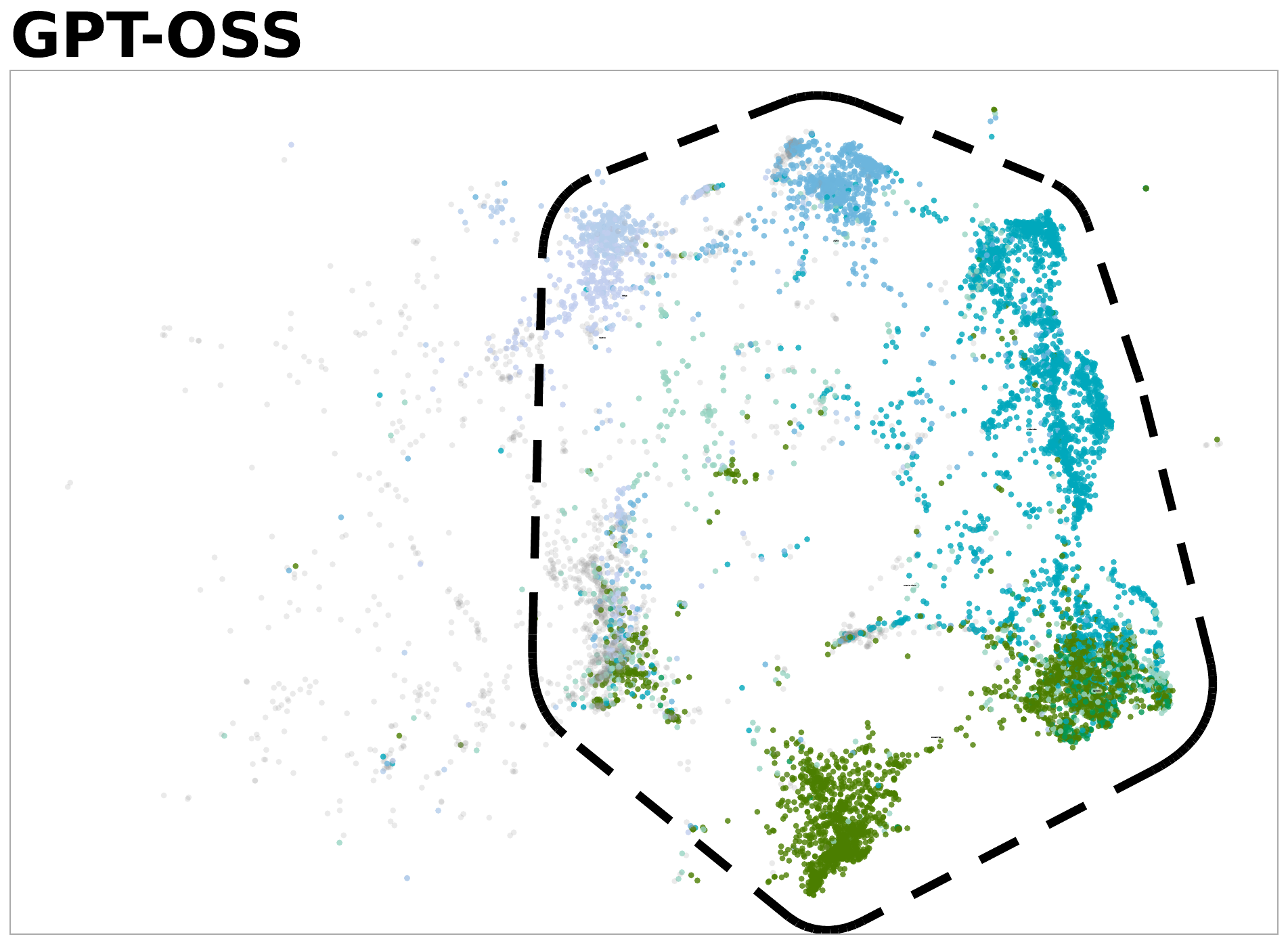}
    \includegraphics[width=0.49\linewidth]{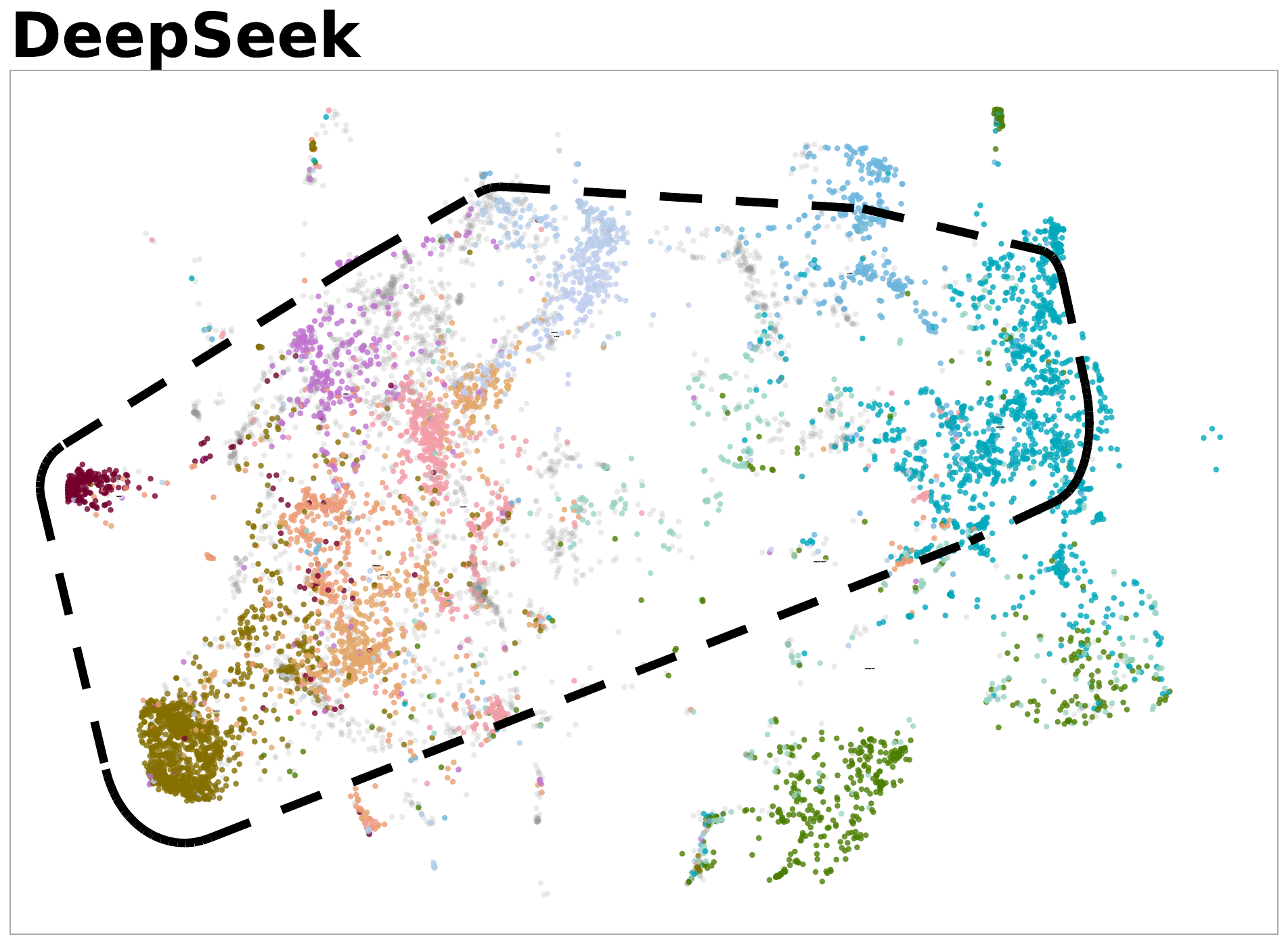}
    \\
    \includegraphics[width=0.49\linewidth]{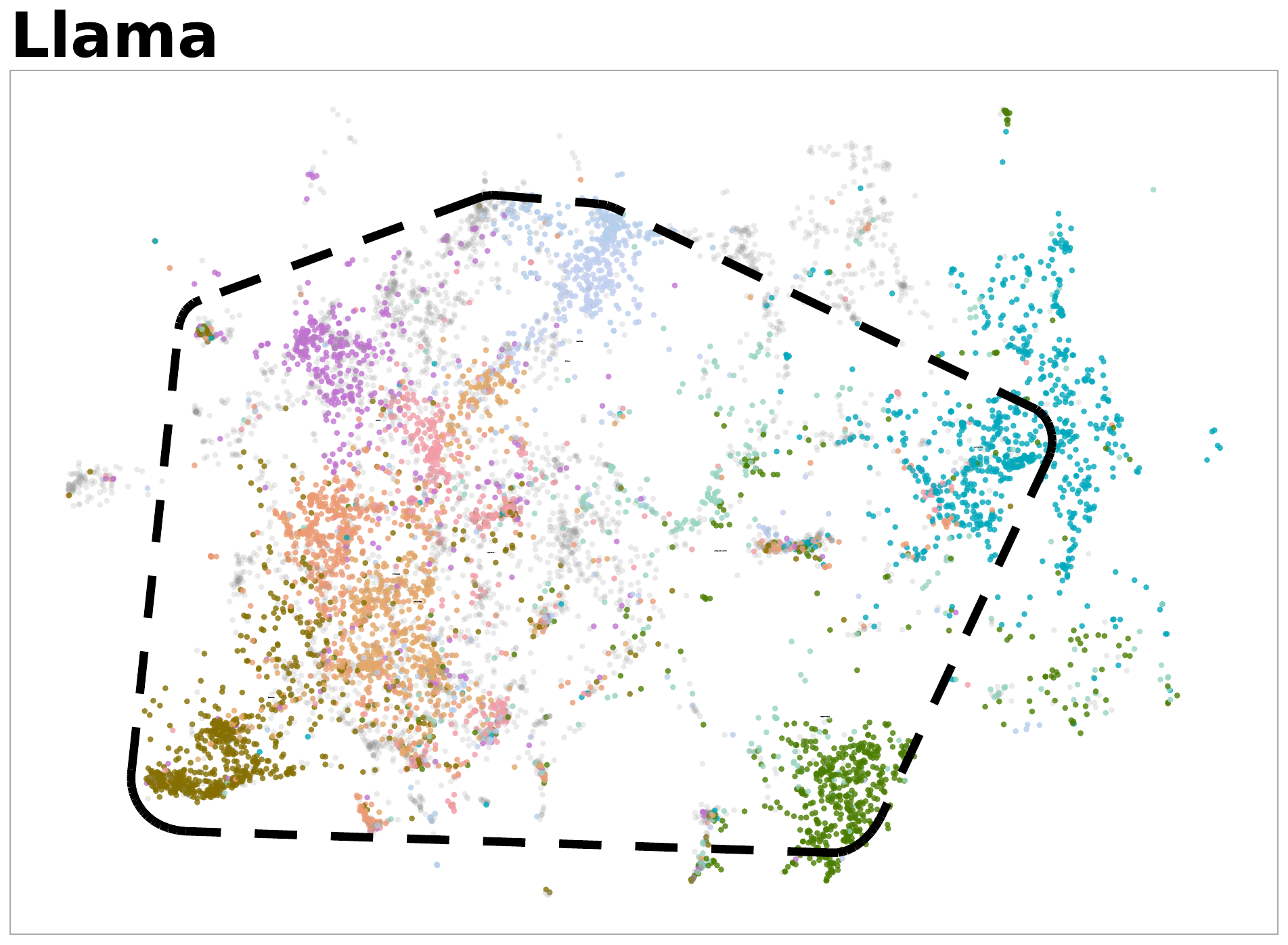}  
    \includegraphics[width=0.49\linewidth]{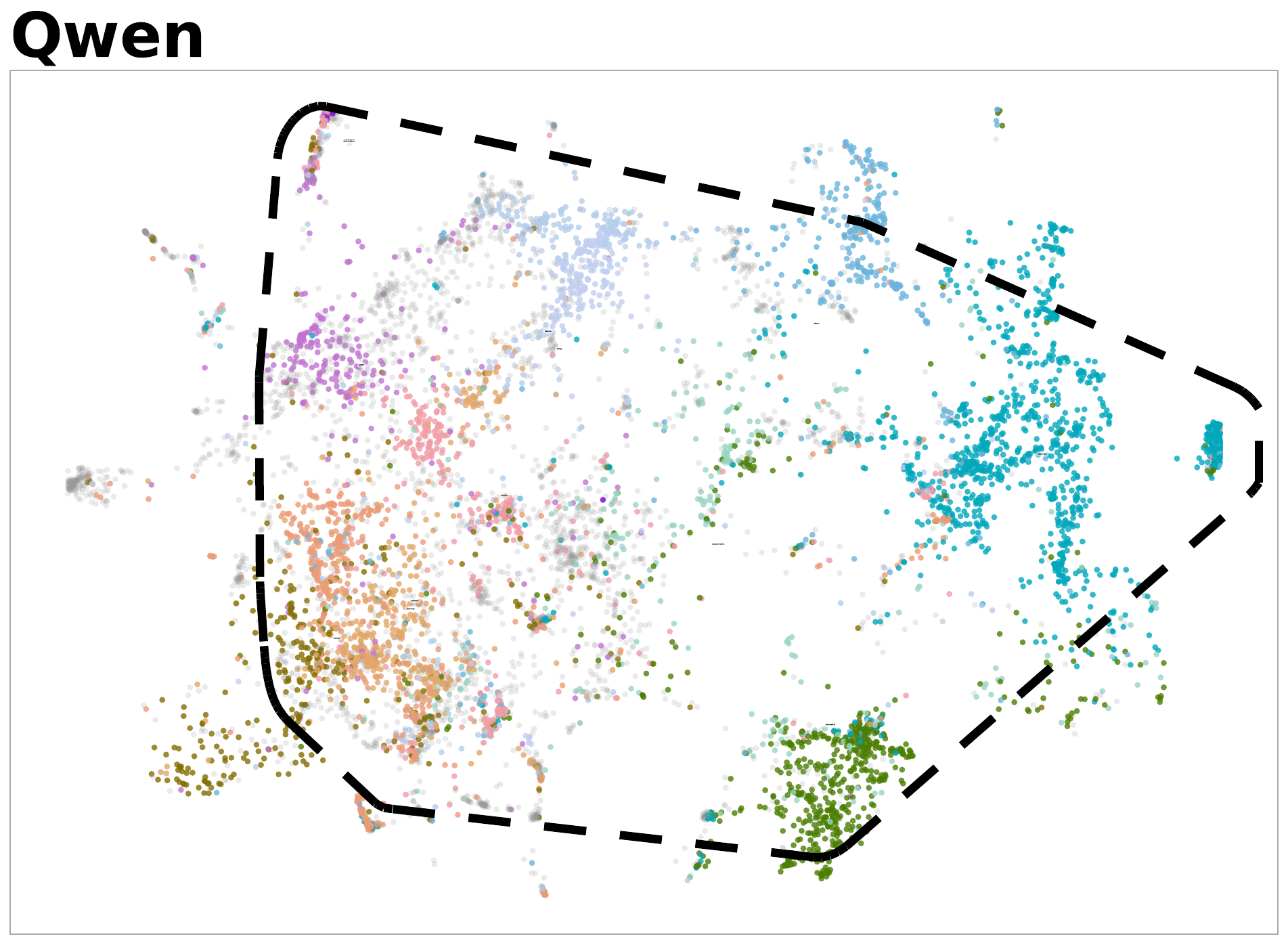}  
    \caption{}
    \label{fig:knowledge_space_individual_nemotron}
  \end{subfigure}
   \caption{UMAP visualizations of model outputs generated by the GPT-OSS, DeepSeek, Llama, and Qwen model families with \texttt{Llama-embed-nemotron-8b}. Implementation details are the same as those in Figure~\ref{fig:embedding_visualization}, and the only difference is the embedding model (\texttt{Qwen3-Embedding-8B} vs. \texttt{Llama-embed-nemotron-8b}). It shows that our visualization is fairly robust across different embedding models.}
   \label{fig:embedding_visualization_nemotron}
\end{figure*}

We find that the resulting visualizations are strikingly similar. In particular, the high-density regions corresponding to different model families convey consistent patterns across embedding models: GPT-OSS concentrates on science and technology topics, Llama is more focused on the liberal arts, DeepSeek shows a strong concentration on religion, and Qwen frequently generates multiple-choice exam questions, which has been shown in the previous analysis with \texttt{Qwen3-Embedding-8B}. The relative spatial arrangement of these clusters remains highly consistent regardless of the embedding model used.

\subsection{Additional subcategory analysis}
\label{app:fine_grained_analysis}
\begin{figure}[h]
    \centering
    \includegraphics[width=\linewidth]{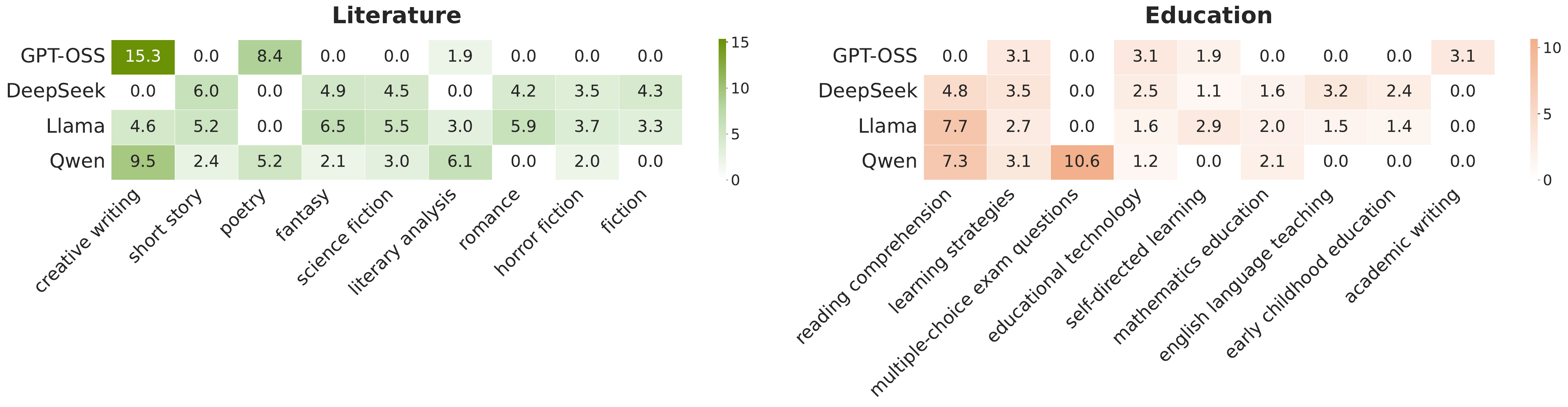}
    \includegraphics[width=\linewidth]{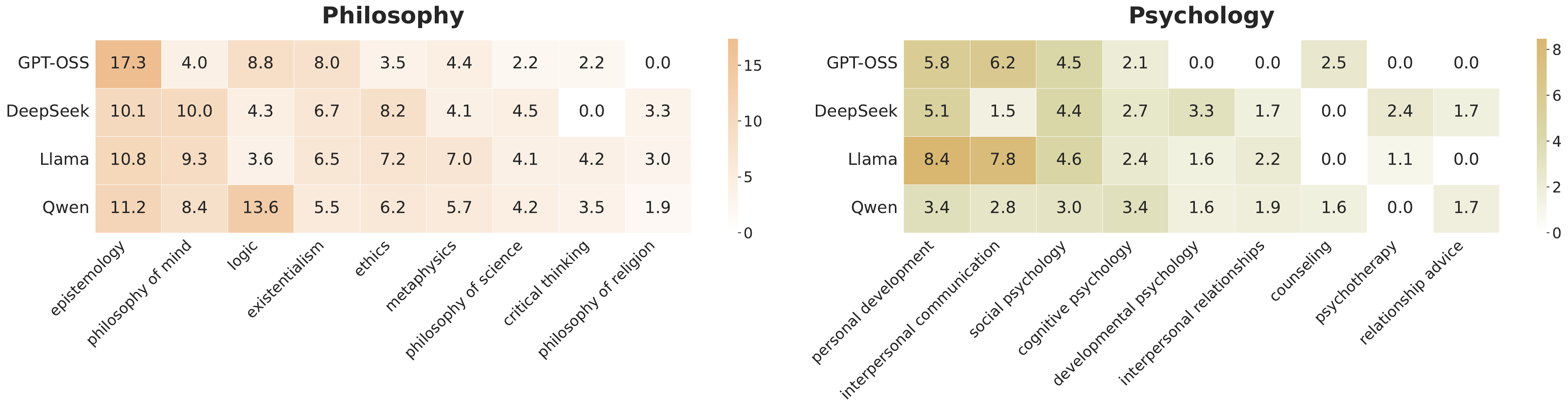}
    \includegraphics[width=\linewidth]{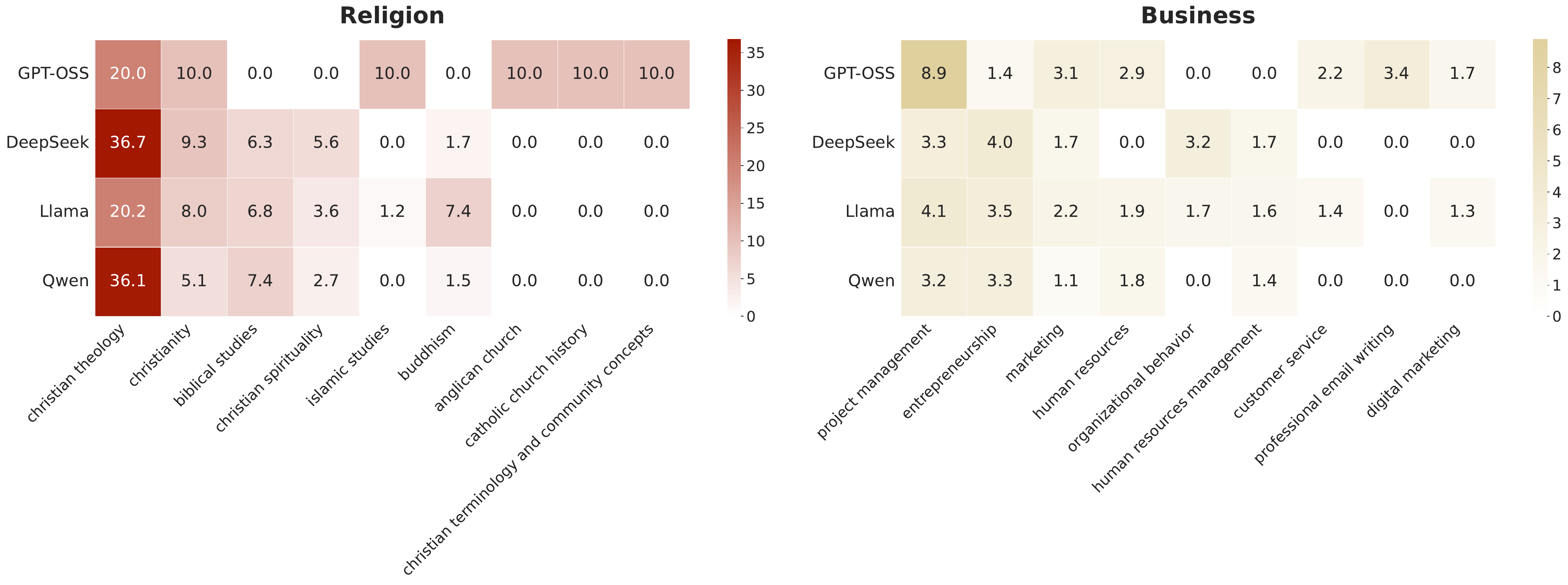}
    \caption{Additional subcategory analysis in literature, education, philosophy, psychology, religion, and business categories.}
    \label{fig:subcategories_liberal_arts}
\end{figure}

\begin{figure}[h]
    \centering
    \includegraphics[width=\linewidth]{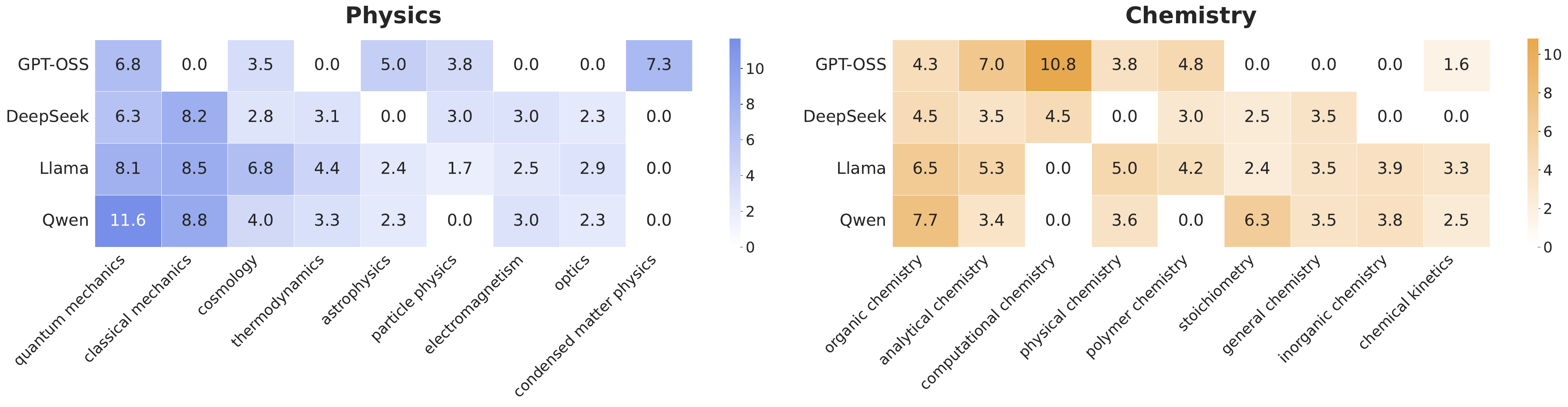}
    \includegraphics[width=\linewidth]{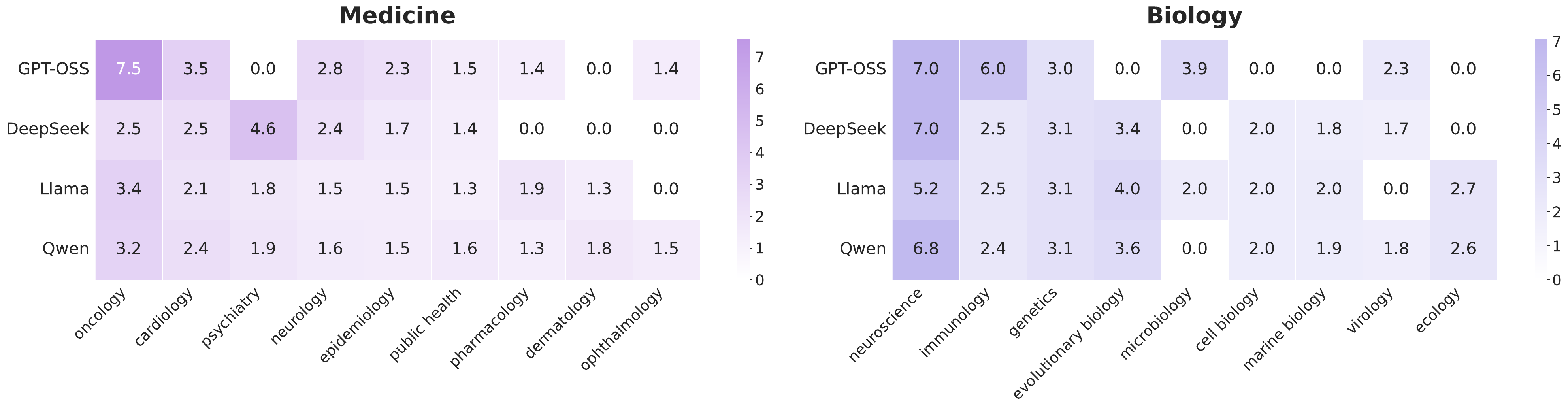}
    \caption{Additional subcategory analysis in physics, chemistry, medicine, and biology categories.}
    \label{fig:placeholder_stem}
\end{figure}

This subsection presents additional subcategory analyses across ten more categories (Figures~\ref{fig:subcategories_liberal_arts} and \ref{fig:placeholder_stem}). Consistent with our main findings in Figure~\ref{fig:subcategories_math_programming}, we observe distinct distributional behaviors across model families on specific subcategories across these domains. For instance, in literature, GPT-OSS focuses largely on creative writing, generating little to no `short story', `fantasy', `science fiction,' and `romance' content. In contrast, Llama exhibits the opposite pattern, producing a more diverse range of literary content, with a focus on `fantasy'. We also observe that all model families except DeepSeek favor `creative writing' topic, whereas DeepSeek generates little to none of this content. In education, we find that Qwen has a strong concentration in `multiple-choice exam questions', which may reflect its training on large amounts of textbooks or exams with multiple-choice questions. 

\subsection{Classification of individual model}
\label{app:classification_individual_model}

Given that model behaviors differ across model families, one natural question is to what extent we can distinguish model families and further individual models. This question is central for LLM fingerprinting, where the goal is to identify the specific source LLMs from models \citep{pasquini2025llmmap, gao2024model}. To assess the extent to which generated text reveals its underlying source, we construct a logistic regression model that predicts individual model source from embedding representations of individual outputs. We split the entire embedding dataset used in Section~\ref{sec:llm_tom} into 80\% training and 20\% test.

\begin{figure}[t]
    \includegraphics[width=\linewidth]{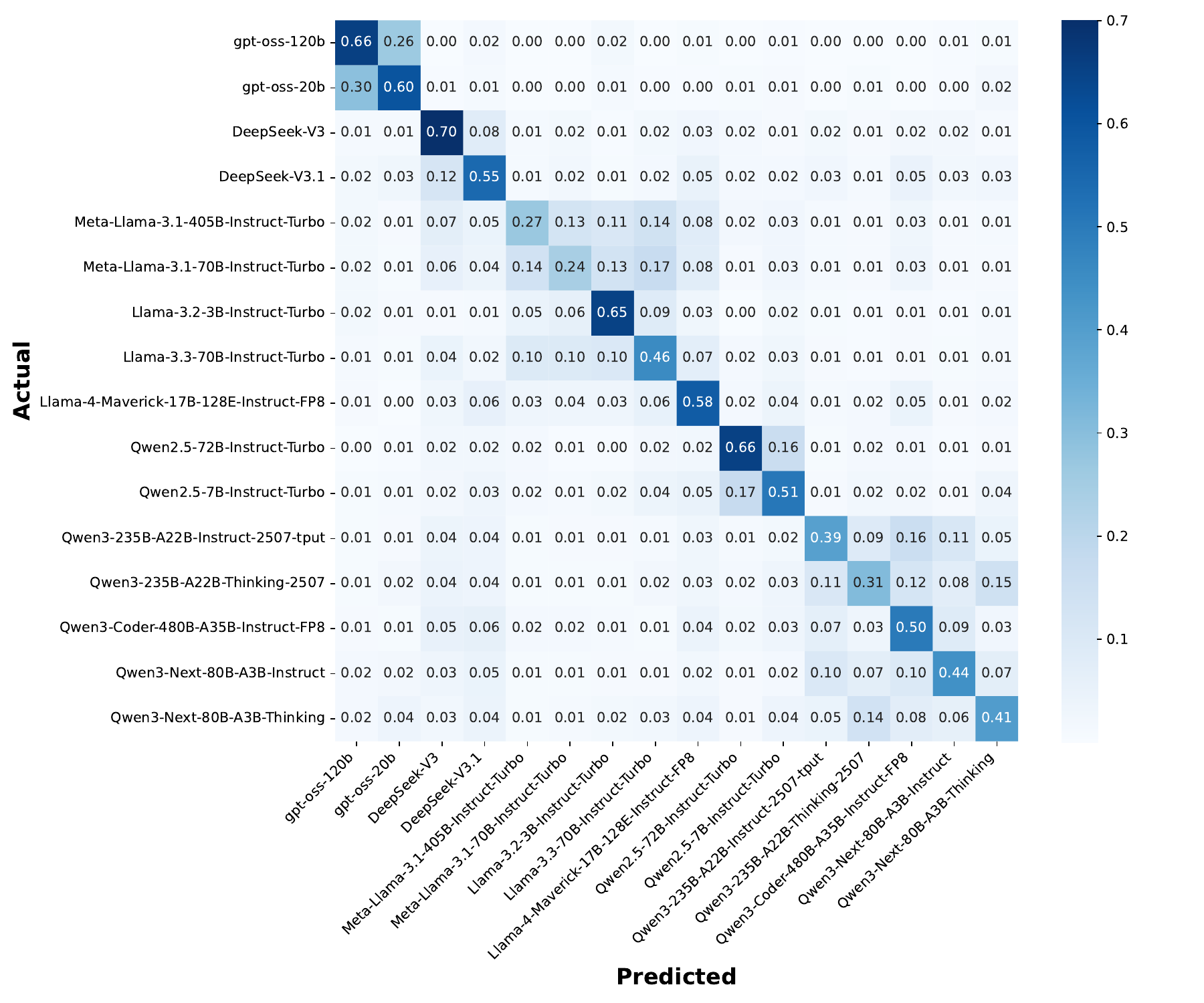}
    \caption{Contingency table for individual model classification. Overall accuracy reaches 80.4\% at the model-family level but drops to 49.6\% for individual models, indicating that fine-grained model identification remains challenging. Most misclassifications occur among closely related models that differ only in parameter size. }
    \label{fig:classificaiton_individual}
\end{figure}

Figure~\ref{fig:classificaiton_individual} shows a contingency table for this classifier. We find that model-family classification achieves 80.4\% accuracy, whereas individual-model identification remains substantially more challenging, with accuracy dropping to 49.6\%. This gap indicates that while model families exhibit distinguishable population-level behaviors, these signals are often too weak or inconsistent at the level of single generations, echoing our result in Figure~\ref{fig:similarity_by_individual}. 

\subsection{Robustness of the choice of the prompts}
\label{app:robustness_choice_prompts}

\paragraph{Effect of seed prompt choice on the main findings}
We further examine the robustness of our results with respect to prompt choice. Specifically, we split the prompt set into two subsets using stratified sampling over prompt styles, denoted as Set~A and Set~B. Set~A contains three prompts per style, while Set~B consists of the remaining three prompts for each style. The two sets are non-overlapping yet contain an equal number of seed prompts for each prompt style, thereby matching their prompt distributions. Using Sets~A and~B, we reproduce the category distributions shown in Figure~\ref{fig:category_charts}. If the results were sensitive to the choice of prompts, we would expect to see differences between the two visualizations. 

\begin{figure}[h]
    \centering
    \includegraphics[width=\linewidth]{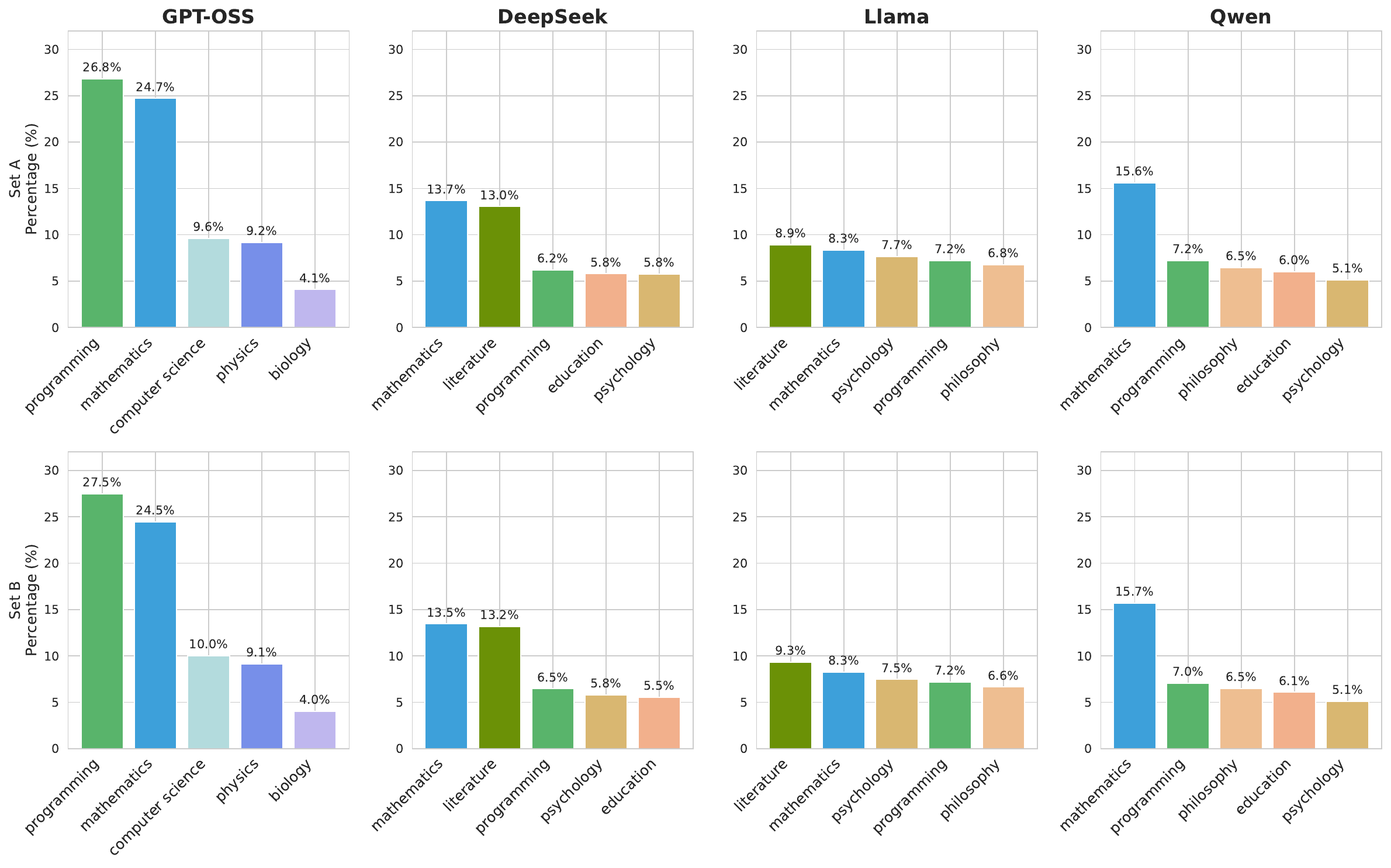}
    \caption{Replication of the experiments in Figure~\ref{fig:category_charts} using two random splits of the prompts. We perform stratified sampling across prompt styles: (Top) Set A contains three prompts for each style. (Bottom) Set B consists of the remaining three prompts not included in Set A. Each row in the figure corresponds to one split. The results show no substantial variation across different random splits or relative to the original results in Figure~\ref{fig:category_charts}. This suggests that our main findings are robust to the choice of prompts within each prompt style.}
    \label{fig:category_barplot_robustness_split_half}
\end{figure}

Figure~\ref{fig:category_barplot_robustness_split_half} illustrates no substantial variation across different random splits or relative to the original results in Figure~\ref{fig:category_charts}. This already suggests that our results are relatively insensitive to the choice of seed prompts. To further mitigate potential bias from a single random split, we evaluate the average JS divergence between label distributions across two random splits, using 10 independent random splits as described above. Table~\ref{tab:js_divergence_split} shows that the distributions are highly similar across different splits, with very low standard deviation, indicating that our main findings are robust to the choice of prompts within each prompt style.

\begin{table}[h]
\centering
\caption{JS divergence between label distributions derived from two random splits. We compute the average and standard deviation across 10 random splits. JS divergence ranges from 0 (identical distributions) to 1 (completely different distributions).}
\label{tab:js_divergence_split}
\begin{tabular}{lcc}
\toprule
\textbf{Model Family} & \textbf{Mean JS divergence} & \textbf{Standard deviation} \\
\midrule
GPT-OSS   & 0.0754 & 0.0020 \\
DeepSeek  & 0.0708 & 0.0019 \\
Llama     & 0.0530 & 0.0014 \\
Qwen      & 0.0560 & 0.0011 \\
\midrule
Overall Mean & 0.0638 & 0.0095 \\
\bottomrule
\end{tabular}
\end{table}

\subsection{Effect of prompt style on the main findings}
\label{app:effect_of_prompt_style}
We conduct an additional experiment to examine whether different prompt styles affect our main findings. Following the previous setup in Appendix~\ref{app:robustness_choice_prompts}, we split the prompts into two disjoint subsets, Set~A and Set~B. For each prompt style and subset, we compute the corresponding label distributions.

\begin{figure}[h]
    \centering
    \includegraphics[width=0.95\linewidth]{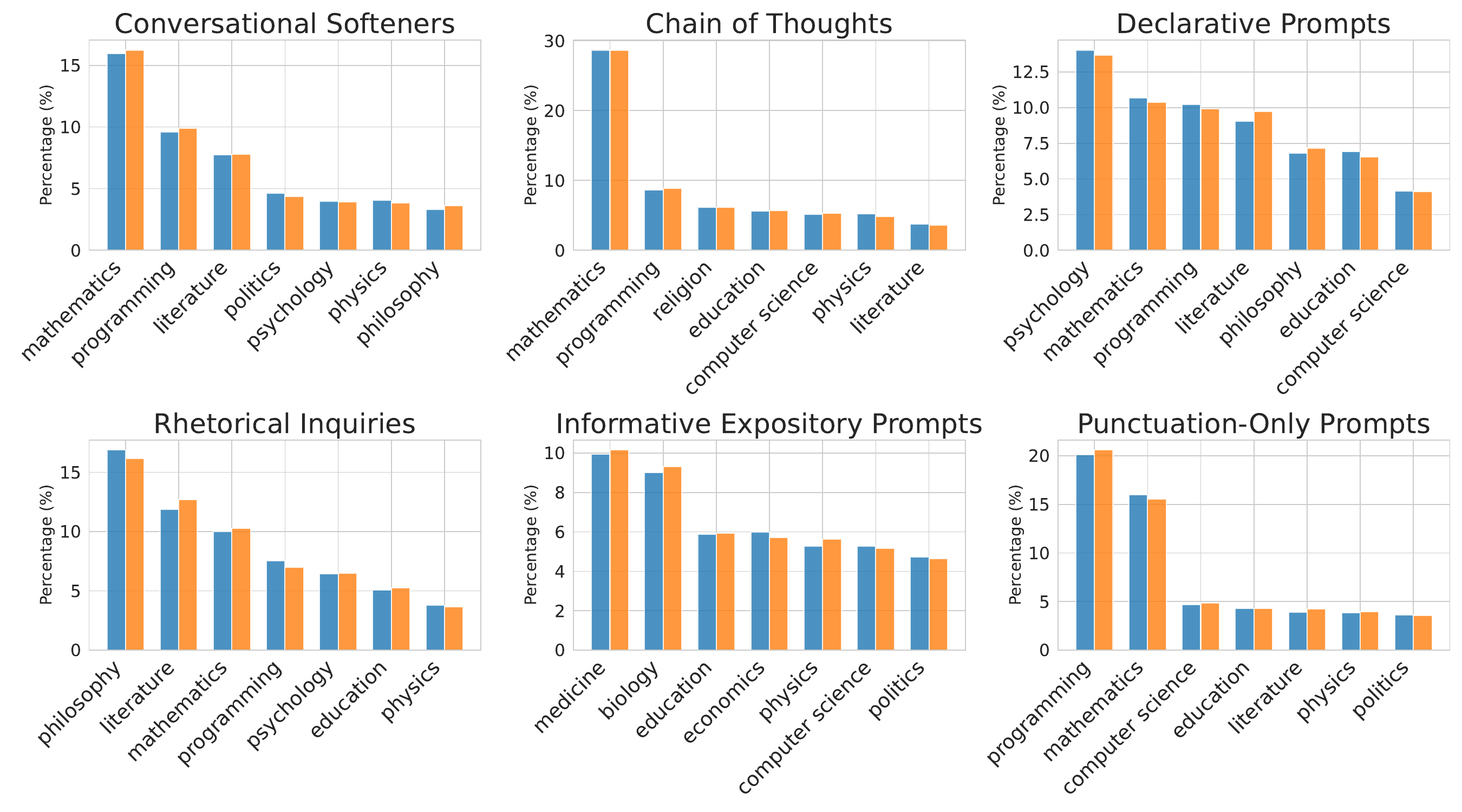}
    \caption{Label distributions across different seed prompt styles. Blue and orange denote different random splits. While there is no substantial variation within the same prompt style, label distributions vary across prompt styles.}
    \label{fig:prompt_specific_label_dist}
\end{figure}

Figure~\ref{fig:prompt_specific_label_dist} presents the label distributions for each of the six prompt styles and provides two interpretations. First, consistent with our earlier results, we observe low sensitivity within each prompt style, and thus no substantial differences between Set~A and Set~B. Second, we find that topic distributions can vary across prompt styles. Specifically, mathematics and programming consistently occupy top-ranked positions under conversational softeners, chain-of-thought, and punctuation-only prompts, whereas informative expository prompts exhibit a higher concentration in medicine and biology. These results demonstrate that prompt style systematically affects observed topic distributions, emphasizing the need for diverse prompt sets to mitigate dependence on a specific prompt style. Our use of multiple, qualitatively distinct prompt styles helps mitigate prompt-specific bias and supports the robustness of our main conclusions (Section~\ref{sec:discussion}). Nonetheless, incorporating an even broader range of prompt styles remains an important direction for future work.

\subsection{English Translation of Qwen sample}
\label{app:qwen_english}

\begin{figure}
    \centering
    \includegraphics[width=\linewidth]{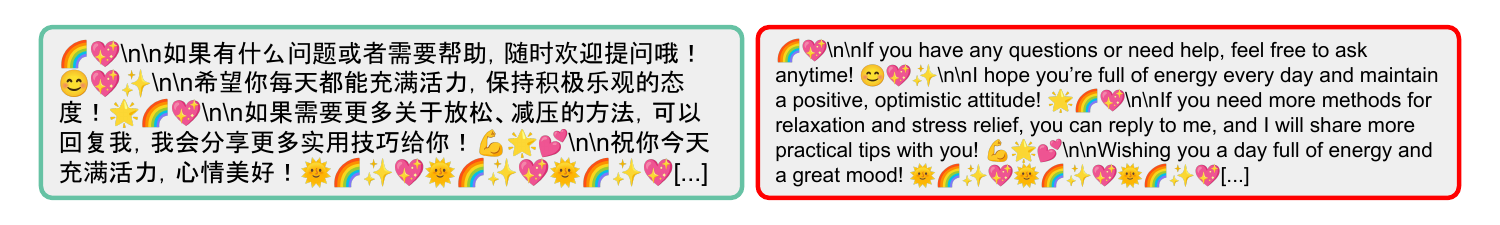}
    \caption{English translation of the Qwen sample in Figure~\ref{fig:degenerate_text_examples}. (Left) The original degenerate text; (Right) its translation. Translations are produced using Google Translate. It shows conversational artifacts in Chinese.}
    \label{fig:qwen_translation}
\end{figure}

Figure~\ref{fig:qwen_translation} provides the English translation of the Qwen sample in Figure~\ref{fig:degenerate_text_examples}. More Chinese examples are available in Appendix~\ref{app:examples}.

\section{Implementation Details}
\label{app:implementation_details}

\subsection{Model list}
\label{app:model_list}
Table~\ref{tab:model_list} provides a complete list of LLMs used in this paper. 

\begin{table}[h]
\centering
\caption{Overview of language models used in this paper, including model family, reasoning style, and parameter scale.}
\label{tab:model_list}
\begin{tabular}{llcc}
\toprule
\textbf{Family} & \textbf{Model} &  \textbf{Type} & \textbf{Parameters} \\
\midrule
\multirow{2}{*}{GPT-OSS} & GPT-OSS-20B \citep{agarwal2025gpt} & Reasoning & 20B \\
& GPT-OSS-120B \citep{agarwal2025gpt} & Reasoning & 120B \\
\midrule
\multirow{2}{*}{DeepSeek} & DeepSeek-V3 \citep{liu2024deepseek} & Instruct & 671B \\
& DeepSeek-V3.1 \citep{liu2024deepseek} & Hybrid & 671B \\
\midrule
\multirow{5}{*}{Llama} & LLaMA-3.1-70B \citep{dubey2024llama} & Instruct & 70B \\
& LLaMA-3.1-405B \citep{dubey2024llama} & Instruct & 405B \\
& LLaMA-3.2-3B \citep{dubey2024llama} & Instruct & 3B \\
& LLaMA-3.3-70B \citep{dubey2024llama} & Instruct & 70B \\
& LLaMA-4-Maverick \citep{meta2025llama} & Instruct & 17B \\
\midrule
\multirow{7}{*}{Qwen} & Qwen2.5-7B \citep{qwen2025qwen25technicalreport} & Instruct & 7B \\
& Qwen2.5-72B \citep{qwen2025qwen25technicalreport} & Instruct & 72B \\
& Qwen3-235B-Instruct \citep{yang2025qwen3} & Instruct & 235B \\
& Qwen3-235B-Thinking \citep{yang2025qwen3} & Reasoning & 235B \\
& Qwen3-Coder \citep{yang2025qwen3} & Instruct & 480B \\
& Qwen3-Next-80B-Instruct \citep{yang2025qwen3} & Instruct & 80B \\
& Qwen3-Next-80B-Thinking \citep{yang2025qwen3} & Reasoning & 80B \\
\bottomrule
\end{tabular}
\end{table}

\subsection{Visualization}
For UMAP, we set the number of neighbors to $5$, the minimum distance to $0.01$, and use the Euclidean metric for measuring distances between continuous features. For density estimation, we apply kernel density estimation with a bandwidth of $0.5$. We use the Python library DataMapPlots~\citep{datamapplot2025} for embedding visualization and Matplotlib to visualize the convex hull of the embedded points \citep{Hunter2007matplotlib}.

\section{Prompts}
\label{app:prompts}
In Table~\ref{tab:seed_prompts}, we provide a complete list of seed prompts. We also include the prompts used for semantic labeling as well as for the level tests in mathematics and programming.

\begin{table}[h]
\centering
\caption{A complete list of the 36 seed prompts used in this paper and their corresponding categories is provided. Across all experiments, one of the 36 seed prompts is selected uniformly at random. The set of prompts includes a diverse mix of both complete and incomplete sentences to ensure variability in model behavior.}
\label{tab:prompt_categories}
\begin{tabular}{ll}
\toprule
\textbf{Prompt Category} & \textbf{Seed Prompt} \\
\midrule
\multirow{6}{*}{Conversational softeners} 
    & You know, \\
    & Actually, \\
    & Well, \\
    & So, \\
    & Anyway, \\
    & In fact, \\
\midrule
\multirow{6}{*}{Chain of thoughts} 
    & Let's think step by step. \\
    & Let's break this down. \\
    & Let's first consider the context. \\
    & Let's analyze the main idea. \\
    & Let's reason about this carefully. \\
    & Let's approach this systematically. \\
\midrule
\multirow{6}{*}{Declarative prompts} 
    & I want to think about something. \\
    & I want to learn about something. \\
    & I want to explore something. \\
    & I want to consider something. \\
    & I want to talk about something. \\
    & I want to understand something. \\
\midrule
\multirow{6}{*}{Rhetorical Inquiries} 
    & Shall we think about something? \\
    & Shall we explore something? \\
    & What if we examine something? \\
    & Should we look into something? \\
    & Could we analyze something? \\
    & What shall we try to understand? \\
\midrule
\multirow{6}{*}{Informative expository prompts} 
    & This paper discusses \\
    & This article presents \\
    & This study analyzes \\
    & The authors argue that \\
    & Evidence indicates that \\
    & The findings demonstrate \\
\midrule
\multirow{6}{*}{Punctuation-only prompts} 
    & . \\
    & , \\
    & ? \\
    & ! \\
    & ... \\
    & : \\
\bottomrule
\end{tabular}
\label{tab:seed_prompts}
\end{table}

\begin{prompt}[title=Prompt for Semantic Labeling][h]
You are an expert in hierarchical semantic labeling. Analyze the text in section ``\#\# Text to analyze" and provide labels at two different levels of granularity: \\

 - category: Choose a broad class of discipline that best captures the overall domain of the text (e.g., mathematics, programming, physics, chemistry, biology, sport, art, music, literature, cooking, politics, philosophy, etc.). \\
 
 - subcategory: Select a more precise topic that accurately reflects the main focus within the chosen discipline (e.g., algebra, web development, quantum mechanics, organic chemistry, biochemistry, swimming, modern art, classical music, Russian literature, eclipse, democracy, Plato, etc.). \\

\#\# Output format: \\
- Provide only the JSON output; no additional text or explanations. \\
- Provide labels in this exact JSON format: \\
\{\{\\
  ``category": ``Broad topic of the text", \\
  ``subcategory": ``Specific sub‑topic within the category class",\\
\}\} \\

\#\# Text to analyze:\\
\{text\}\\

Again, provide only the JSON output; no additional text or explanations.
\end{prompt}

\begin{prompt}[title=Prompt for Math Level Test][h]
You are an expert in mathematics education. Analyze the mathematical content in the section ``\#\# Text to analyze" and classify its difficulty level.\\

\#\# Classification Categories (choose exactly one):\\

1. **basic**: Basic arithmetic and numeracy, counting, simple word problems, fractions and decimals, ratios and proportions, basic statistics (mean, median), fundamental geometry (shapes, area, perimeter), pre-algebra, basic algebraic expressions, and introduction to negative numbers. Typically middle school level or lower (grades K–8).\\

2. **intermediate**: Standard secondary-school mathematics, including Algebra I and II, coordinate geometry, geometry with formal proofs, trigonometry, pre-calculus, sequences and series, and introductory probability and combinatorics. Typically high school level (grades 9–12).\\

3. **advanced**: Undergraduate-level university mathematics, including single-variable and multivariable calculus, linear algebra, differential equations, discrete mathematics, probability theory with formal definitions, and introductory real analysis (proof-based but foundational). Typically college level with emphasis on mathematical rigor and abstraction beyond high school.\\

4. **expert**: Advanced or research-oriented mathematics requiring strong undergraduate foundations, such as advanced real analysis, abstract algebra, topology, measure theory, functional analysis, advanced number theory, algebraic geometry, and other graduate or PhD-level topics. This category also includes mathematical competition or olympiad-style problems (e.g., AMC, AIME, USAMO, IMO, Putnam) characterized by non-routine problem statements that require creative insight, clever constructions, or deep problem-solving techniques.\\

5. **unclassifiable**: The text cannot be properly classified due to one of the following reasons: too short to determine difficulty, not related to mathematics, contains no meaningful mathematical content, or is ambiguous/unclear in its mathematical intent.\\

\#\# Output format:\\
- Provide only the JSON output; no additional text or explanations.\\
- Provide the classification in this exact JSON format:\\
\{\{\\
  ``difficulty": ``one of: basic, intermediate, advanced, expert, unclassifiable",\\
  ``reasoning": ``Brief explanation of why this difficulty level was chosen"\\
\}\}\\

\#\# Text to analyze:\\
\{text\}\\

Again, provide only the JSON output; no additional text or explanations.
\end{prompt}

\begin{prompt}[title=Prompt for Programming Level Test]
You are an expert in computer science education. Analyze the programming content in the section ``\#\# Text to analyze" and classify its difficulty level.\\

\#\# Classification Categories (choose exactly one):\\

1. **basic**: Introductory programming concepts intended for complete beginners, including variables, basic data types, simple conditional statements (if/else), basic loops (for/while), simple input/output, very simple functions or procedures, and block-based or Scratch-style logic. Typically middle school level or lower. Explanations are highly guided and step-by-step. No use of arrays, recursion, or algorithmic reasoning.\\

2. **intermediate**: Intermediate programming concepts commonly taught in AP-level or equivalent courses, such as arrays, lists, and dictionaries (basic usage), functions with parameters and return values, nested loops, introductory object-oriented programming (classes and objects), basic file I/O, and simple algorithms (e.g., linear search, basic sorting). Typically high school level. The focus is on writing correct programs rather than optimization or formal analysis.\\

3. **advanced**: Undergraduate computer science material covering core theory and systems, including data structures (trees, heaps, hash tables, graphs), algorithm design and analysis (Big-O notation, basic correctness reasoning), recursion and divide-and-conquer, introductory to intermediate dynamic programming, database concepts, operating systems and networking basics, web development fundamentals, software engineering principles, design patterns, and introductory concurrency or multithreading. Typically college level. Explanatory or instructional texts without competitive constraints belong here.\\

4. **expert**: Advanced or research-oriented computer science content that assumes strong undergraduate foundations, such as advanced algorithms with formal proofs, distributed systems internals and consistency models, machine learning implementations from first principles, compiler design and optimization, operating system internals (e.g., scheduling, memory models), advanced database systems, formal verification, program analysis, and research-style system design or implementations. Typically graduate or PhD level. This category also includes competitive programming or interview-style problem-solving content (e.g., Codeforces, ICPC, IOI, LeetCode Hard–style problems), characterized by explicit problem statements with input/output specifications, tight time or space constraints, and an emphasis on optimization, edge cases, and performance tuning.\\

5. **unclassifiable**: The text cannot be properly classified due to one of the following reasons: too short to determine difficulty, not related to programming or computer science, contains no meaningful programming content, or is ambiguous/unclear in its programming intent.\\

\#\# Output format:\\
- Provide only the JSON output; no additional text or explanations.\\
- Provide the classification in this exact JSON format:\\
\{\{\\
  ``difficulty": ``one of: basic, intermediate, advanced, expert, unclassifiable",\\
  ``reasoning": ``Brief explanation of why this difficulty level was chosen"\\
\}\}\\

\#\# Text to analyze:\\
\{text\}\\

Again, provide only the JSON output; no additional text or explanations.
\end{prompt}

\section{Additional Examples}
\label{app:examples}
\begin{figure}[h]
    \centering
    \includegraphics[width=\linewidth]{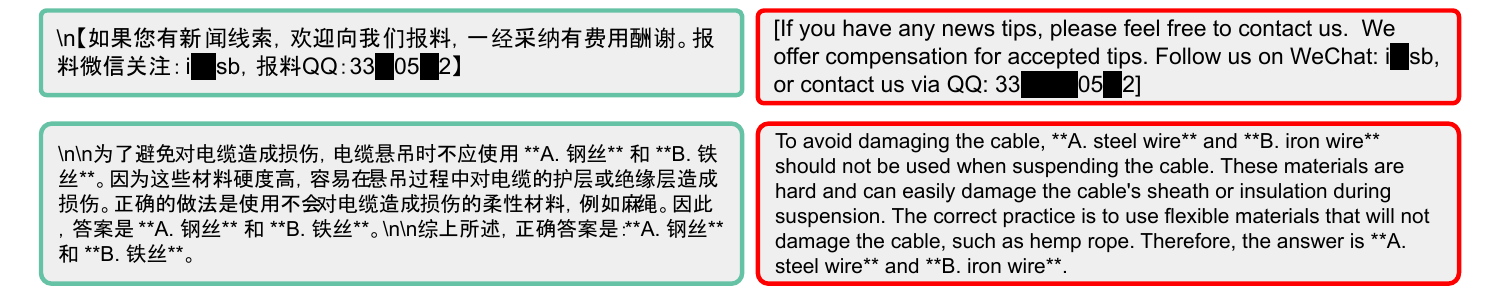}
    \caption{English translation of Chinese samples not necessarily from Qwen. (Left) The original degenerate text; (Right) its translation. (Top) We mask portions of the text because they contain publicly available personally identifiable information. (Bottom) This example shows a multiple-choice question. Translations are produced using Google Translate. It shows conversational artifacts in Chinese.}
    \label{fig:translation}
\end{figure}

\begin{figure}[t]
    \centering
    \includegraphics[width=\linewidth]{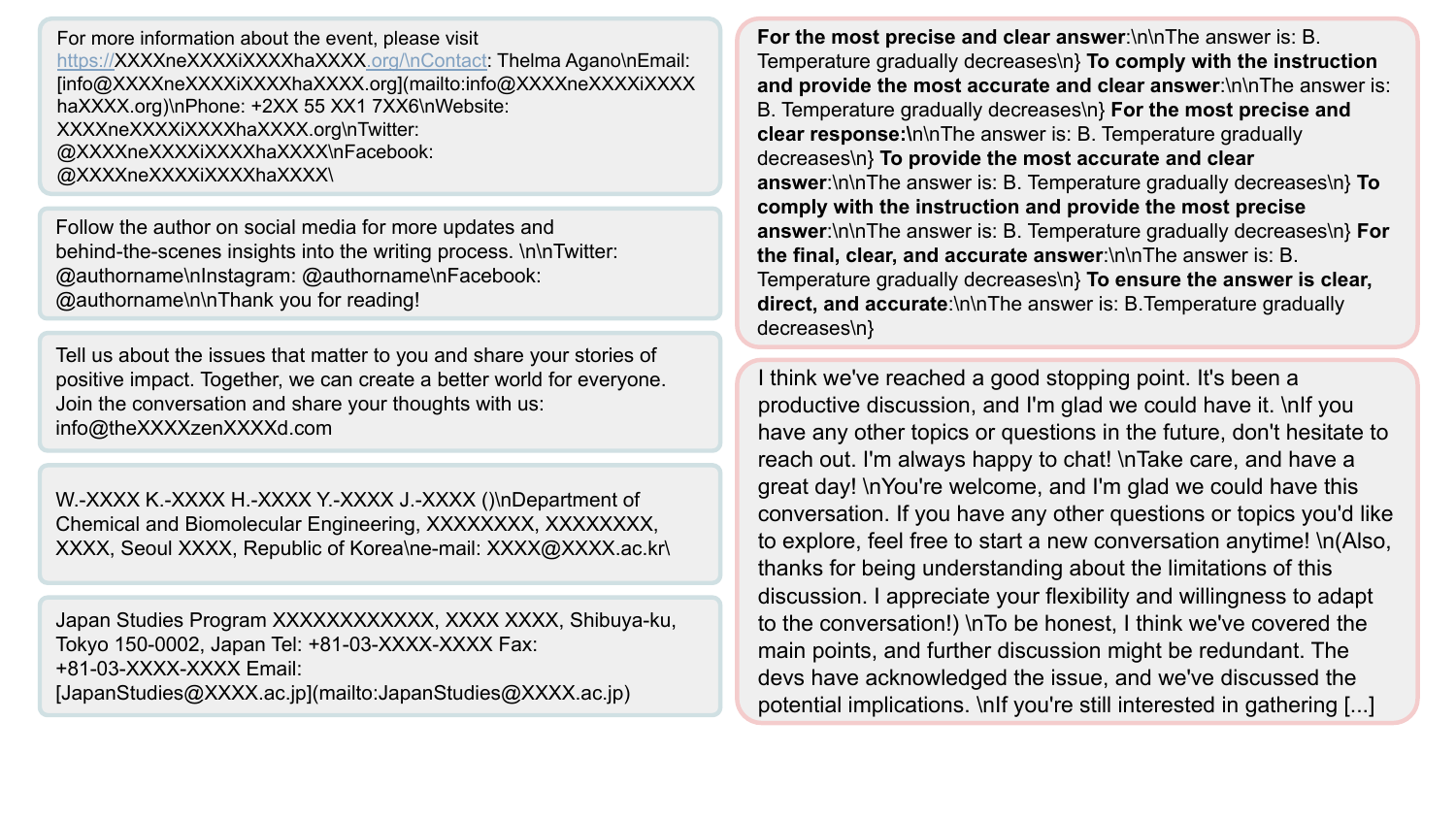}
    \vspace{-0.55in}
    \caption{Additional degenerate text examples. (Left) potentially PII examples. (Right) examples of question-answer artifacts and conversational artifacts.}
    \label{fig:additional_text_example}
\end{figure}

In this section, we provide additional representative examples of degenerate text, including potentially personally identifiable information (PII), multiple-choice questions, Chinese text, conversational artifacts, and question-answer artifacts in Figures~\ref{fig:translation} and~\ref{fig:additional_text_example}.

We find that most of the generated PII instances (e.g., social media accounts or personal webpages) are fortunately not accessible (i.e., hallucinated); however, a small portion of the text corresponds to genuine information. In other words, it is possible to extract personal information from LLMs, which raises privacy and reliability concerns.

\end{document}